\documentclass[10pt,twocolumn,letterpaper]{article}

\usepackage{cvpr}
\usepackage{times}
\usepackage{epsfig}
\usepackage{graphicx}
\usepackage{amsmath}
\usepackage{amssymb}

\usepackage{amsthm}
\usepackage{algorithm}
\usepackage{algorithmic}
\usepackage{subfigure}
\usepackage{epstopdf}
\usepackage{xr}
\usepackage[page]{appendix}
\usepackage{enumitem}
\usepackage[pagebackref=true,breaklinks=true,letterpaper=true,colorlinks,bookmarks=false]{hyperref}

 \cvprfinalcopy % *** Uncomment this line for the final submission

\def\cvprPaperID{1258} % *** Enter the CVPR Paper ID here

\def\ie{i.e.~}
\def\eg{e.g.~}

\def\st{~~\textrm{s.t.}~~}
\def\and{\textrm{and}}
\def\conv{\textrm{conv}}

\def\b{\textbf{b}}
\def\0{\textbf{0}}
\def\1{\textbf{1}}

\def\r{\boldsymbol{r}}

\def\x{\boldsymbol{x}}

\def \bfdelta{\boldsymbol{\delta}}
\def \bfmu{\boldsymbol{\mu}}
\def \bfpi{\boldsymbol{\pi}}
\def \bff{\boldsymbol{f}}

\def\I{\mathbf{I}}
\def\II{\mathbb{I}}
\def \bfdelta{\boldsymbol{\delta}}
\def \bfpi{\boldsymbol{\pi}}

\def\E{\mathcal{E}}

\def\I{\mathcal{I}}

\def\P{\mathcal{P}}

\def\S{\mathcal{S}}

\def\transpose{\top}

\newtheorem{assumption}{Assumption}[section]
\newtheorem{definition}{Definition}[section]
\newtheorem{lemma}{Lemma}[section]

\newtheorem{theorem}{Theorem}[section]
\newtheorem{corollary}{Corollary}[section]
\newtheorem{problem}{Problem}[section]

\newtheorem*{lemma*}{Lemma}
\newtheorem*{theorem*}{Theorem}

\newcommand{\RR}{I\!\!R} %real numbers
\newcommand{\myparagraph}[1]{\smallskip\noindent\textbf{#1.}}

\ifcvprfinal\fi
\begin{document}

%%%%%%%%% TITLE
\title{Sparse Representation Graph for Outlier Detection in Subspaces }
\title{Provable Self-Representation Based Outlier Detection in a Union of Subspaces}
%\title{Provable Self-Representation Based Outlier Detection for Subspace Clustering}

\author{Chong You, Daniel P. Robinson, Ren\'e Vidal\\
	Johns Hopkins University, Baltimore, MD, 21218, USA\\
	%Institution1 address\\
	%	{\tt\small \{cyou,rvidal\}@cis.jhu.edu}
	% For a paper whose authors are all at the same institution,
	% omit the following lines up until the closing ``}''.
	% Additional authors and addresses can be added with ``\and'',
	% just like the second author.
	% To save space, use either the email address or home page, not both
	%\and
	%Second Author\\
	%Institution2\\
	%First line of institution2 address\\
	%{\tt\small secondauthor@i2.org}
}

\maketitle
\thispagestyle{empty}

%%%%%%%%% ABSTRACT
\begin{abstract}
	Many computer vision tasks involve processing large amounts of data contaminated by outliers, which need to be detected and rejected. While outlier detection methods based on robust statistics have existed for decades, only recently have methods based on sparse and low-rank representation been developed along with guarantees of correct outlier detection when the inliers lie in one or more low-dimensional subspaces. This paper proposes a new outlier detection method that combines tools from sparse representation with random walks on a graph. By exploiting the property that data points can be expressed as sparse linear combinations of each other, we obtain an asymmetric affinity matrix among data points, which we use to construct a weighted directed graph. By defining a suitable Markov Chain from this graph, we establish a connection between inliers/outliers and essential/inessential states of the Markov chain, which allows us to detect outliers by using random walks. We provide a theoretical analysis that justifies the correctness of our method under geometric and connectivity assumptions. Experimental results on image databases demonstrate its superiority with respect to state-of-the-art sparse and low-rank outlier detection methods.
	%submitted to CVPR:
	%Many computer vision tasks involve processing large amounts of data contaminated by outliers, which need to be detected and rejected. While outlier detection methods based on robust statistics have existed for decades, it is only recently that methods based on sparse and low-rank representation have been developed and that guarantees of correct outlier detection have been established when the inliers lie in one or more low-dimensional subspaces. This paper proposes a new outlier detection method that combines tools from sparse representation theory and graph theory. By exploiting the property that data points can be expressed as linear combinations of each other, we obtain an asymmetric affinity matrix among data points, which we use to construct a weighted directed graph. By defining a suitable random walk on this graph, we establish a connection between inliers/outliers and essential/inessential states of a Markov chain, which allows us to directly detect outliers. We also provide a theoretical analysis that justifies the correctness of our method under geometric and connectivity assumptions. Experimental results on image databases demonstrate its superiority with respect to state-of-the-art sparse and low rank outlier detection methods.
\end{abstract}
\vspace{-1em}
%%%%%%%%% BODY TEXT
\section{Introduction}
\label{sec:intro}

%Subspace and low-dimensional structures associated with data arise in many computer vision applications such as motion segmentation \cite{Kanatani:ICCV01} and face classification \cite{Basri:PAMI03}. A lot of research activity has been dedicated to the analysis of such data in topics such as subspace learning and subspace clustering. Since data collection is now-a-days often automated, points that do not lie in the subspaces, \ie outliers, are frequently obtained. It is often essential to detect and reject these outliers before any subsequent processing/analysis is performed.
In many applications in computer vision, including motion estimation and segmentation \cite{Kanatani:ICCV01} and face recognition \cite{Basri:PAMI03}, high-dimensional datasets can be well approximated by a union of low-dimensional subspaces. Such applications have motivated a lot of research on the problems of learning one or more subspaces from data, a.k.a.  subspace learning and subspace clustering, respectively. In practice, datasets are often contaminated by points that do not lie in the subspaces, \ie outliers. In such situations, it is often essential to detect and reject these outliers before any subsequent processing/analysis is performed.

\myparagraph{Prior work} 
We address the problem of outlier detection in the setting when the inlier data are assumed to lie close to a union of unknown low-dimensional subspaces (low relative to the dimension of the ambient space).
A traditional method for solving this problem is RANSAC \cite{RANSAC}, which is based on randomly selecting a subset of points, fitting a subspace to them, and counting the number of points that are well fit by this subspace; this process is repeated for sufficiently many trials and the best fit is chosen. RANSAC is intrinsically combinatorial and the number of trials needed to find a good estimate of the subspace grows exponentially with the subspace dimension. Consequently, the methods of choice have been to robustly learn the subspaces by penalizing the sum of \emph{unsquared} distances (in lieu of \emph{squared} distances used in classical methods such as PCA) of points to the closest subspace \cite{Ding:ICML06,Lerman:AS11,Zhang:IJCV12,Zhang:WSM09}. Such a penalty is robust to outliers because it reduces the contributions from large residuals arising from outliers. However, the optimization problem is usually nonconvex and a good initialization is extremely important for finding the optimal solution. 
% The following: CVPR submitted
%For the past decade, the methods of choice have been to seek a low rank reconstruction residual by using outlier-robust penalties~\cite{Ding:ICML06,Zhang:IJCV12}. For example, the method R1-PCA proposed by \cite{Ding:ICML06} seeks a factorization of $X$, which contains data in its columns, into a subspace basis $U$ and coefficient matrix $V$ by solving the problem
%\begin{equation}
%	\min_{U, V} \|X - UV^\top\|_{2,1} \st U^\transpose U = I,
%	\label{eq:R1_PCA}
%\end{equation}
%where $\|A\|_{2,1}$ is the sum of the $\ell_2$ norms of the columns of the matrix $A$. The model \eqref{eq:R1_PCA} is an adaptation of canonical PCA in which the sum of \emph{unsquared} penalties is used in place of the squared penalties, thus reducing the contributions from large residuals arising from columns of $X$ that are outliers. Unfortunately, there is no known closed form solution to the nonconvex problem \eqref{eq:R1_PCA}, so that an alternating minimization approach is usually adopted, which is typically sensitive to initialization. Moreover, there are no theoretical guarantees for the correctness of the method.
%%Consequently, a good initialization is extremely important for finding the globally optimal subspace. 

\begin{figure*}[h]
	\centering
	\subfigure[\label{fig:SelfRepresentation_1} Two exemplar representation vectors]{\includegraphics[scale = 0.82,trim={7cm 4.4cm  15.5cm 8.5cm},clip]{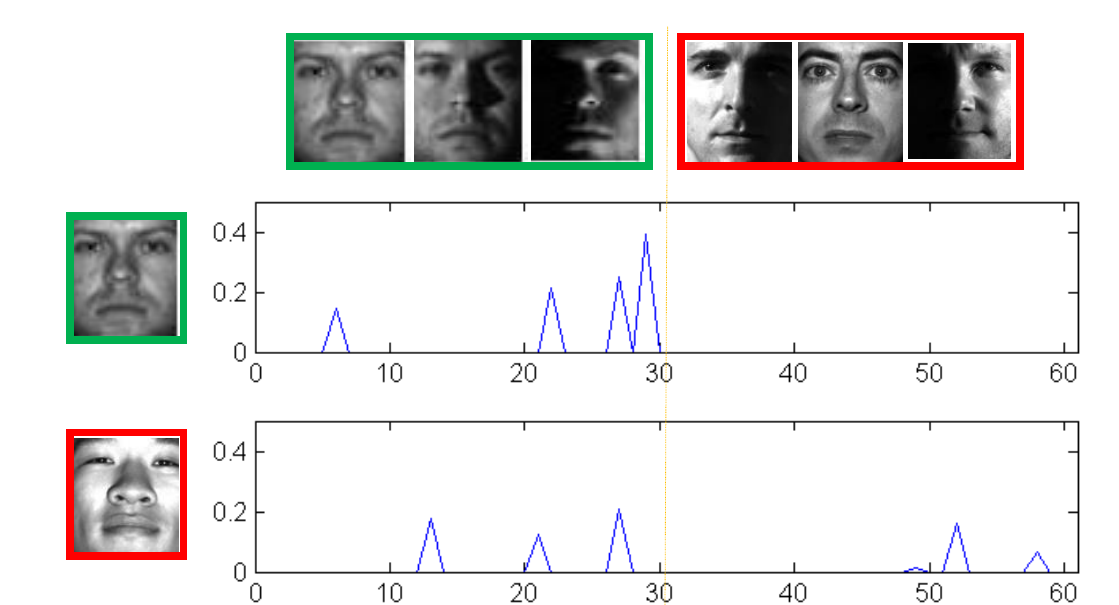}}
	%	\subfigure[\label{fig:SelfRepresentation_1} Two exemplar representation vectors]{\includegraphics[scale = 0.36,trim={6cm 3cm  6cm 3cm},clip]{SelfRepresentation_1_newnew}}
	~~~~~~~
	%\subfigure[\label{fig:SelfRepresentation_2} Representation matrix $R$]{\includegraphics[scale = 0.5,trim={4cm 9cm  5cm 10cm},clip]{SelfRepresentation_2_new}}
	\subfigure[\label{fig:SelfRepresentation_2} Representation matrix $R$]{\includegraphics[scale = 0.3,trim={6cm 1.3cm  6cm 1cm},clip]{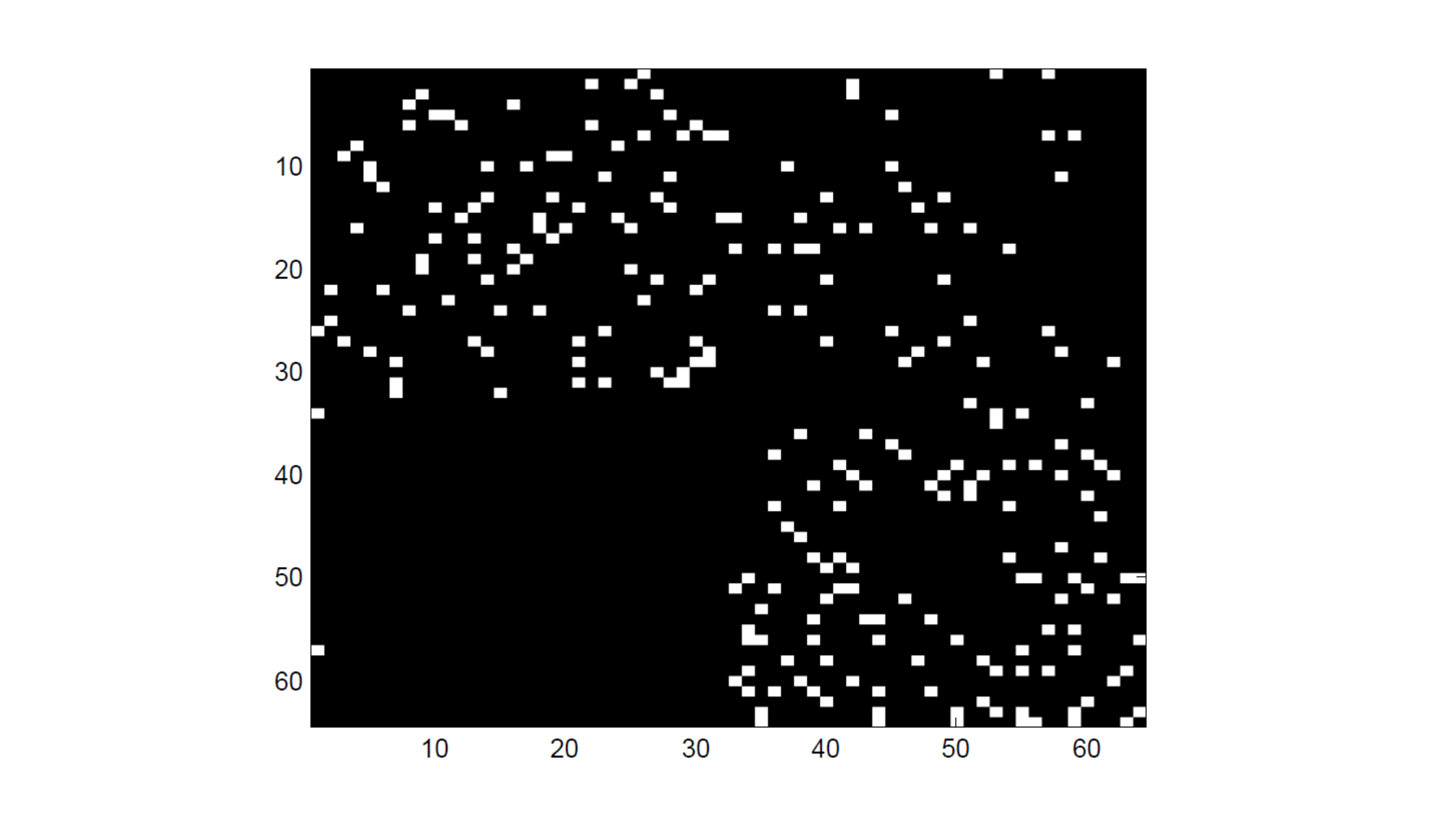}}
	\caption{An illustration of a self-representation matrix $R$ in the presence of outliers. The first $32$ columns of the data matrix $X$ correspond to $32$ images of one individual under different illuminations from the Extended Yale B database, and the next $32$ images are randomly chosen from all other individuals; three examples from each category are shown near the top of \ref{fig:SelfRepresentation_1}. We also show a typical representation vector for an inlier and an outlier image in \ref{fig:SelfRepresentation_1}, and the complete  representation matrix $R$ in \ref{fig:SelfRepresentation_2}, where white and black denote $r_{ij} \neq 0$ and $r_{ij}=0$. Notice that inliers use only other inliers in their representation, while outliers use both inliers and outliers in their representations.}
	\label{fig:SelfRepresentation}
	\vspace{-0.5em}
\end{figure*} 

The groundbreaking work of Wright et al. \cite{Wright2009a} and Cand\`es et al. \cite{Candes:ACM10} on using convex optimization techniques to solve the PCA problem with robustness to corrupted entries has led to many recent methods for PCA with robustness to outliers \cite{Xu:NIPS10,Mccoy:EJS11,Li:TSP15-outlier,Zhang:JMLR14,Lerman:FCM15}. For example, Outlier Pursuit \cite{Xu:NIPS10} uses the nuclear norm $\|\cdot\|_*$ to seek low-rank solutions by solving the problem   $\min_{L} \|X - L\|_{2,1} + \lambda \|L\|_*$
for some $\lambda>0$. 
%REAPER \cite{Lerman:FCM15} models the subspace by the orthoprojector $\Pi$ that minimizes $\|X - \Pi X\|_{2,1}$ and relaxes the orthoprojection constraint to a convex constraint.
A prominent advantage of convex optimization techniques is that they are guaranteed to correctly identify outliers under certain conditions. Very recently, several nonconvex outlier detection methods have also been developed with guaranteed correctness \cite{Lerman:Arxiv14,Cherapanamjeri:Arxiv17}. Nonetheless, these methods typically model a \emph{unique} inlier subspace, e.g., by a low rank matrix $L$ in Outlier Pursuit, and therefore cannot deal with multiple inlier subspaces since the union of multiple subspaces could be high-dimensional. 

Another class of methods with theoretical guarantees for correctness utilizes the fact that outliers are expected to have low similarities with other data points. In~\cite{Chen:IJCV09,ArChLe:EJStat11}, a multi-way similarity is introduced that is defined from the polar curvature, which has the advantage of exploiting the subspace structure. However, the number of combinations in multi-way similarity can be prohibitively large. Some recent works have explored using inner products between data points for outlier detection \cite{Heckel:TIT15,Rahmani:arXiv16}. %For example, the coherence pursuit (CoP) \cite{Rahmani:arXiv16} claims a point to be an outlier if the sum of its inner products with all other points is small. 
Although computationally very efficient, these methods require the inliers to be well distributed and densely sampled within the subspaces.

%Motivated by the observation that data in high-dimensional ambient space (outliers) are sparse relative to data in low-dimensional subspaces (inliers), several recent methods address outlier detection by computing correlation between data points. In \cite{Heckel:TIT15}, a point is categorized as an outlier if its inner products with all other points are small. The coherence pursuit (CoP) algorithm \cite{Rahmani:arXiv16} claims a point to be an outlier if the sum of its inner products with all other points is small. Although these methods can deal with multiple inlier subspaces, they require the inliers to be well distributed and densely sampled within the subspaces.

\myparagraph{Overview of our method and contributions} In this work, we address the problem of outlier detection by using data self-representation. The proposed approach builds on the self-expressiveness property of data in a union of low-dimensional subspaces, originally introduced in \cite{Elhamifar:CVPR09}, which states that a point in a subspace can always be expressed as a linear combination of other points in the subspace. In particular, if the columns of $X=[\x_1, \cdots, \x_N]$ lie in multiple subspaces, then for all $j=1,\dots, N$, there exists a vector $\r_j \in \RR^N$ such that $\x_j = X \r_j$ and the nonzero entries of $\r_j$ correspond to points in the same subspace as $\x_j$. If the subspace dimensions are small, $\r_j$ can be taken to be sparse and be computed by solving the $\ell_1$ minimization problem
\begin{equation}
	\min_{\r_j} \|\r_j\|_1 + \frac{\gamma}{2} \|\x_j - X \r_j\|_2^2 ~~~ \text{s.t.} ~ r_{jj} = 0
	\label{eq:selfrepresentation}
\end{equation}
for some $\gamma > 0$.
In \cite{Elhamifar:CVPR09}, an \emph{undirected} graph is constructed from $R = [\r_1, \cdots, \r_N]$ in which each vertex corresponds to a data point, and vertices corresponding to $\x_i$ and $\x_j$ are connected if either $r_{ij}$ or $r_{ji}$ is nonzero. Such a graph can be used to segment the data into their respective subspaces by applying spectral clustering \cite{vonLuxburg:StatComp2007} to the graph's Laplacian.

Consider now the case where $X$ contains outliers to the subspaces. Figure~\ref{fig:SelfRepresentation} illustrates an example representation matrix $R$ computed from \eqref{eq:selfrepresentation} for data drawn from a single subspace (face images from one individual) plus outliers (other images). In this case, the representation $R$ is such that inliers express themselves as linear combinations of a few other inliers, while outliers express themselves as linear combinations of both inliers and outliers.  Motivated by this observation, we use a \emph{directed} graph to model data relations: a directed edge from $\x_j$ to $\x_i$ indicates that $\x_j$ uses $\x_i$ in its representation (i.e. $r_{ij}\ne 0$). Then a random walk on the representation graph initialized at an outlier will not return to the set of outliers since once the random walk reaches an inlier it cannot return to the outliers. Therefore, we design a random walk process and identify outliers as those whose probabilities tend to zero. Our work makes the following contributions with respect to the state of the art:
\begin{enumerate}[topsep=0.25em,noitemsep]
\item Our method can detect outliers using the probability distribution of a \emph{random walk} on a graph constructed from \emph{data self-representation}.
\item Our \emph{data self-representation} allows our method to handle multiple inlier subspaces. Knowledge of the number of subspaces and their dimensions is not required, and the subspaces may have a nontrivial intersection.
\item Our method can explore contextual information by using a \emph{random walk}, i.e., the ``outlierness'' of a particular point depends on the ``outlierness'' of its neighbors.
%The stationary probability of a point $\x_j$ combines information from the entire representation $R$, thus is expected to be a more robust characterization of the property of $\x_j$ than, e.g. $\|\r_j\|$ in \cite{Soltanolkotabi:AS12}. 
\item Our analysis shows that our method correctly identifies outliers under suitable assumptions on the data distribution and connectivity of the representation graph. 
\item Experiments on real image databases illustrate the effectiveness of our method. 
\end{enumerate}

\section{Related work} 
\label{sec:prior}

\myparagraph{Outlier detection by self-representation} Prior work has explored using data self-representation as a tool for outlier detection in a union of subspaces. Specifically, motivated by the observation that outliers do not have \emph{sparse} representations, \cite{Soltanolkotabi:AS12,Cong:CVPR11} declare a point $\x_j$ as an outlier if $\|\r_j\|_1$ is above a threshold. However, this $\ell_1$-thresholding strategy is not robust to outliers that are close to each other since their representation vectors may have small $\ell_1$-norms. The LRR \cite{Liu:ICML10} solves for a low-rank self-representation matrix $R$ in lieu of a sparse representation and penalizes the sum of unsquared self-representation errors $\|\x_j - X \r_j\|_2$, which makes it more robust to outliers. However, LRR requires the subspaces to be independent and the sum of the union of subspaces to be low-dimensional~\cite{Liu:AISTAT12}.

\myparagraph{Outlier detection by maximum consensus} In a diverse range of contexts such as maximum consensus \cite{Zheng:CVPR11,Chin:CVPR16} and robust linear regression \cite{Mitra:TSP13,Wang:CVPR15-Regression}, people have studied problems of the form
\begin{equation}	\label{eq:prior-affine}
\min_{\b} \sum_{i=1}^N \II(|\x_i^\transpose \b - y_i| \ge \epsilon),
\end{equation}
in which $\II(\cdot)$ is the indicator function. Note that if we set $y_i = 1$ for all $i$, then \eqref{eq:prior-affine} can be interpreted as detecting outliers in data $X$ where the inliers lie close to an \emph{affine} hyperplane. A problem closely related to \eqref{eq:prior-affine} is 
\begin{equation}\label{eq:prior-linear}
\min_{\b} \sum_{i=1}^N \II(|\x_i^\transpose \b| \ge \epsilon) \st \b \ne 0,
\end{equation}
which appears in many applications (\eg see \cite{Qu:NIPS14}). In particular, \eqref{eq:prior-linear} can be used to learn a \emph{linear} hyperplane from data corrupted by outliers. To detect outliers in a general low-dimensional subspace, one can apply \eqref{eq:prior-affine} and \eqref{eq:prior-linear} recursively to find a basis for the orthogonal complement of the subspace \cite{Tsakiris:DPCPICCV15}. However, such an approach is limited because there can be only one inlier subspace and the dimension of that subspace must be known in advance.

\myparagraph{Outlier detection by random walk} Perhaps the most well-known random walk based algorithm is PageRank \cite{Brin:98pagerank}. Originally introduced to determine the authority of website pages from web graphs, PageRank and its variants have been used in different contexts for ranking the centrality of the vertices in a graph. In particular, \cite{Moonesinghe:ICTAI06,Moonesinghe:IJAIT08} propose the OutRank, which ranks the ``outlierness'' of points in a dataset by applying PageRank to an undirected graph in which the weight of an edge is the cosine similarity or RBF similarity between the two connected data points. Then, points that have low centrality are regarded as outliers. The outliers returned by OutRank are those that have low similarity to other data points. Therefore, OutRank does not work if points in a subspace are not dense enough.

%CVPR submitted version
%Our method is closely related to the spectral ranking methods \cite{Vigna:arXiv15,Mahoney:arXiv16} such as PageRank \cite{Brin:98pagerank} and VisualRank \cite{Jing:PAMI08}. The main differences are that the edges of the graph in ranking mark the endorsement or approval of the predecessor (tail) to the successor (head), and that the random walker finds the most important nodes recursively by using this information. In contrast, the edges in our method do not give endorsement measures. Instead, they represent the different behaviors of inliers and outliers with respect to their connections to other points, which is then combined with random walks to predict the sets of inliers and outliers.

\section{Outlier detection by self-representation}
\label{sec:method}

In this section, we present our data self-representation based outlier detection method. We first describe the data self-representation and its associated properties for inliers and outliers. We then design a random walk algorithm on the representation graph whose limiting behavior allows us to identify the sets of inliers and  outliers.

\subsection{Data self-representation}

Given an unlabeled dataset $X = [\x_1, \cdots, \x_N]$ containing inliers and outliers, the first step of our algorithm is to construct the data self-representation matrix denoted by $R=[\r_1, \cdots, \r_N]$. As briefly discussed in the introduction (see also Figure~\ref{fig:SelfRepresentation}), a self-representation matrix $R$ computed from~\eqref{eq:selfrepresentation} is observed to have different properties for inliers and outliers. Specifically, inliers usually use only other inliers for self-representation, \ie for an inlier $\x_j$, the representation is such that $r_{ij} \ne 0$ only if $\x_i$ is also an inlier, where $r_{ij}$ is the $(i, j)$-th entry of $R$. This property is expected to hold if the inliers lie in a union of low dimensional subspaces, as evidenced from the works \cite{Elhamifar:TPAMI13,Soltanolkotabi:AS12,You:ICML15,Wang:JMLR16,Wang:ICML15}. As an intuitive explanation, if the inliers lie in a low dimensional subspace, then any inlier has a \emph{sparse} representation using other points in this subspace.
%, \eg by using those that form a basis of the subspace. 
Thus such a representation can be found by using sparsity-inducing regularization as seen in~\eqref{eq:selfrepresentation}.  In contrast, outliers are generally randomly distributed in the ambient space, so that 
%ere is no special sparse representations and 
a self-representation usually contains both inliers and outliers.

Since the representation $R$ computed from~\eqref{eq:selfrepresentation} is sparse, there are potentially connectivity issues in the representation graph, \ie an inlier that is not well-connected to other inliers may be detected as an outlier, and an outlier that is not well connected may be detected as an inlier. To address the connectivity issue, we compute the data self-representation matrix $R$ by  the elastic net problem~\cite{Zou:JRSS05,You:CVPR16-EnSC}:
\begin{equation}
\min_{\r_j} \lambda \|\r_j\|_1 + \frac{1-\lambda}{2}  \|\r_j\|_2^2  + \frac{\gamma}{2} \|\x_j - X \r_j \|_2^2  ~~~\text{s.t.}~ r_{jj} = 0,
\label{eq:representation_matrix}
\end{equation}
in which $\lambda \in [0, 1]$ controls the balance between sparseness (via $\ell_1$ regularization) and connectivity (via $\ell_2$ regularization). Specifically, if $\lambda$ is chosen close to $1$, we can still expect that the computed representation for an inlier will only use inliers. The $\ell_2$ regularization has been introduced to promote more connections between data points, \ie if $\lambda\in[0,1)$, then one expects more nonzero entries in $R$. A detailed discussion of the representation computed from~\eqref{eq:representation_matrix} and the connectivity issue is provided in Section \ref{sec:theory}.

\subsection{Representation graph and random walk}
We use a directed graph $G$, which we call a \emph{representation graph}, to capture the behavior of inliers and outliers from the representation matrix $R$. The vertices of $G$ correspond to the data points $X$, and the edges are given by the (weighted) adjacency matrix $A := |R|^\transpose \in \RR^{N \times N}$ with the absolute value taken elementwise,
%, which is defined to be $|R|^\transpose$ ($|\cdot|$ applies to each entry of the matrix),
i.e., the weight of the edge from $\x_i$ to $\x_j$ is given by $a_{ij}=|r_{ji}|$. In the representation graph, we expect that vertices corresponding to  inliers will have edges that only lead to inliers, while vertices that are outliers will have edges that lead to both inliers and outliers. In other words, we do not expect to have any edges that lead from an inlier to an outlier. 

Using the previous paragraph as motivation, we design a random walk procedure to identify the outliers. A random walk on the representation graph $G$ is a discrete time Markov chain, for which the  transition probability from $\x_i$ at a given time to $\x_j$ at the next time is given by $p_{ij} := a_{ij} / d_{i}$ with  $d_{i} := \sum_j a_{ij}$. By this definition, if the starting point of a random walk is an inlier then it will never escape the set of inliers as there is no edge going from any inlier to any outlier. In contrast, a random walk starting from an outlier will likely end up in an inlier state since once it enters any inlier it will never return to an outlier state.  Thus, by using different data points to initialize random walks,  outliers can be identified by observing the final probability distribution of the state of the random walks (see Figure \ref{fig:graph}).

If $P \in \RR ^{N \times N}$ is the transition matrix with entries $p_{ij}$, then $P$ is related to the representation matrix $R$ by
\begin{equation}
p_{ij} = |r_{ji}| / \|\r_i\|_1 \ \ \text{for all } \ \ \{i, j\} \subset \{1, 2, \cdots N\}.
\label{eq:def-P}
\end{equation} 
We define $\bfpi^{(t)} = [\pi_1^{(t)}, \dots, \pi_N^{(t)}]$ to be the state probability distribution at time $t$, then the state transition is given by $\bfpi^{(t+1)} = \bfpi^{(t)}  P$. Thus, a $t$-step transition is $\bfpi^{(t)} = \bfpi^{(0)} P^t$ with $\bfpi^{(0)}$ the chosen initial state probability distribution.

\subsection{Main algorithm: Outlier detection by R-graph}
%Based on the discussion above, 
We propose to perform outlier detection by using random walks on the representation graph~$G$. We set the initial probability distribution as $\bfpi^{(0)}= [1/N, \cdots, 1/N]$, and then compute the $t$-step transition %to get the probability distribution after $t$ steps, \ie 
$\bfpi^{(t)} = \bfpi^{(0)} P^t$. This can be interpreted as initializing a random walk from each of the $N$ data points, and then finding the sum of probability distributions of all random walks after $t$ steps. It is expected that all random walks---starting from either an inlier or an outlier---will eventually have high probabilities for the inlier states and low probabilities for the outlier states. 
%Thereafter, the outliers can be identified as the entries of $\bfpi_t$ that are less than a certain threshold.

\begin{figure}[t]
	\centering
	\includegraphics[scale = 0.53,trim={5.5cm 0.5cm 14cm 7cm},clip]{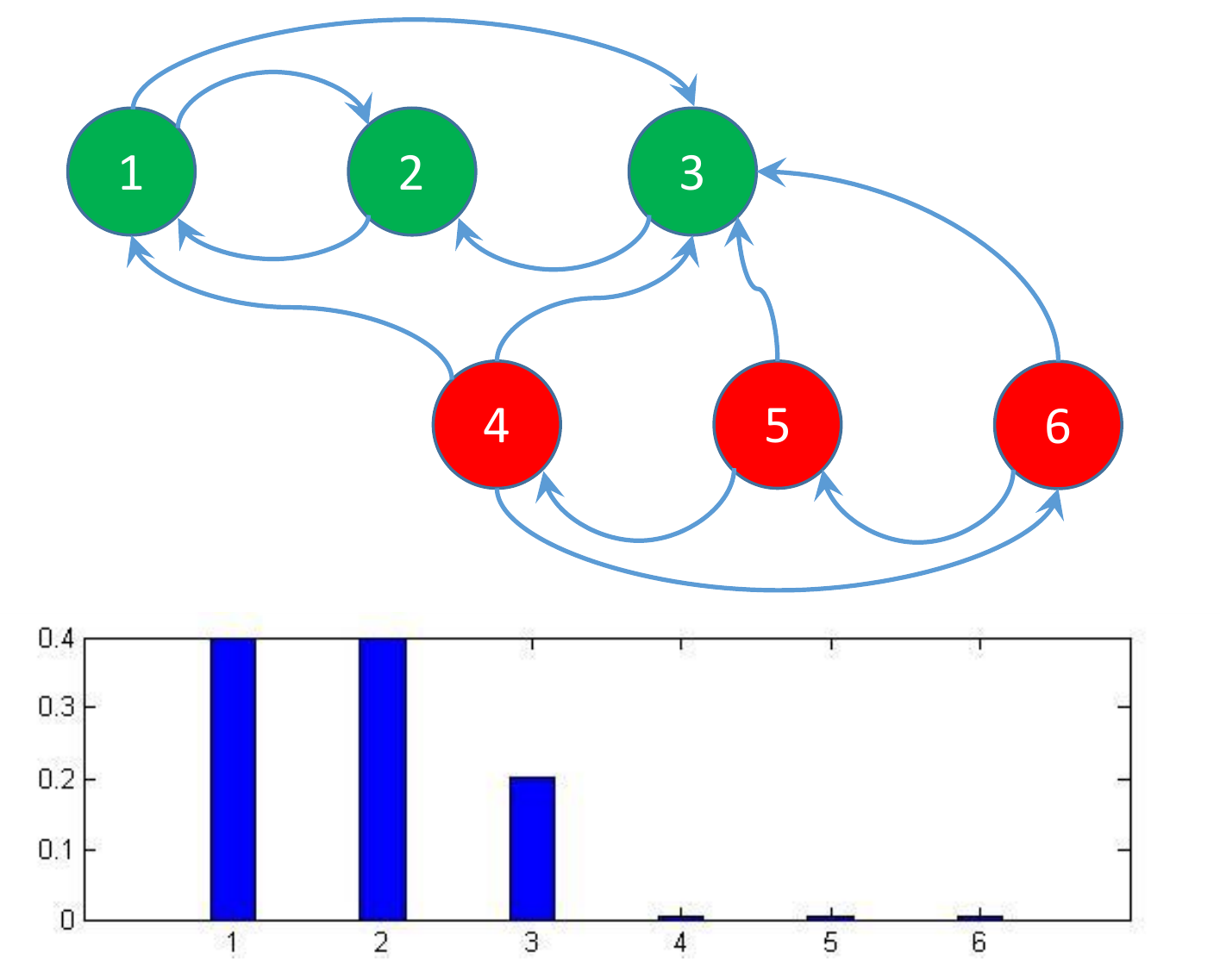}
	\caption{Illustration of random walks on a representation graph. Top: green balls represent inliers and red balls represent outliers, and arrows represent edges among nodes. Notice that there is no edge going from inliers to outliers. A random walk starting from any point will end up at only inlier points. Bottom: bar plot of $\bar{\bfpi}^{(100)}$ with the $i$th bar corresponding to the $i$th entry in $\bar{\bfpi}^{(100)}$. 
The use of thresholding on this probability distribution will correctly distinguish outliers from inliers.}
	\label{fig:graph}
\end{figure}

We note that the $\bfpi^{(t)}$ defined as above need not converge, as shown by the $2$-dimensional example $P = [\begin{smallmatrix} 0&1\\ 1&0 \end{smallmatrix}]$. Instead, we choose to use the $T$-step Ces\`aro mean, given by
\begin{equation}
\bar{\bfpi}^{(T)} 
= \frac{1}{T} \sum_{t=1}^T \bfpi^{(0)} P^t
\equiv \frac{1}{T} \sum_{t=1}^T \bfpi^{(t)},
\label{eq:def-I}
\end{equation}
which is the average of the first $T$ t-step probability distributions (see Figure~\ref{fig:graph}). The sequence $\{\bar{\bfpi}^{(T)}\}$ has the benefit that it always converges, and its limit is the same as that of $\bfpi^{(t)}$ whenever the latter exists. In the next section, we give a more detailed discussion of this choice, its properties for outlier detection, and its convergence behavior.

Our complete algorithm is stated as Algorithm~\ref{alg:main}. 
%Note that $\bar{\bfpi}^{(T)} $ is computed recursively for efficiency.

\begin{algorithm}
	\caption{Outlier detection by representation graph}
	\label{alg:main}
	\begin{algorithmic}[1]
		\REQUIRE Data $X = [\x_1, \cdots\!,\x_N]$, $\#$iterations $T$, threshold $\epsilon$.\!\!\!
		\STATE Use $X$ to solve for $R = [\r_1, \cdots, \r_N]$ using  \eqref{eq:representation_matrix}.
		\STATE \label{step:P}Compute $P$ from $R$ using \eqref{eq:def-P}. 
		\STATE Initialize $t = 0$, $\bfpi = [1/N, \cdots, 1/N]$, and $\bar{\bfpi} = \0$.
		%\LOOP
		\FOR{$t = 1, 2, \dots T$} 
		\STATE Compute $\bfpi \leftarrow \bfpi \cdot P$, and then set $\bar{\bfpi} \leftarrow \bar{\bfpi} + \bfpi$.
		%\STATE If $T=t$, terminate; otherwise set $t\leftarrow t+1$.
		\ENDFOR
		%\ENDLOOP
		\STATE $\bar{\bfpi} \leftarrow \bar{\bfpi} / T$.
		\ENSURE An indicator of outliers: $\x_j$ is an outlier if $\bar{\bfpi}_j \le \epsilon$.
	\end{algorithmic}
\end{algorithm}

\section{Theoretical guarantees for correctness}
\label{sec:theory}

Let us first formally define the problem of outlier detection when data is drawn from a union of subspaces.
\begin{problem}[Outlier detection in a union of subspaces]
	Given data $X =[\x_1, \cdots, \x_N]\in \RR^{D\times N}$ whose columns contain inliers that are drawn from an unknown number of unknown subspaces $\{\S_\ell\}_{\ell=1}^n$, and outliers that are outside of $\cup_{\ell=1}^n \S_\ell$, the goal is to identify the set of outliers.
	\label{pr:outlier-detection}
\end{problem}

Recall that motivation for our method is that ideally there will be no edge going from an inlier to an outlier in the representation graph. This motivates us to assume that a random walk starting at any inlier will eventually return to itself, \ie inliers are \emph{essential states} of the Markov chain, while outliers are those that have a chance of never coming back to itself, \ie outliers are \emph{inessential states}. Formally, we work with a (time homogeneous) Markov chain with state space $\Omega=\{1, \cdots, N\}$, in which each state $j$ corresponds to data $\x_j$, and the transition probability $P$ is given by \eqref{eq:def-P}. Given $\{i,j\}\subset \Omega$, we say that $j$ is accessible from $i$, denoted as $i \to j$, if there exists some $t > 0$ such that the $(i,j)$-th entry of $P^t$ is positive. Intuitively, $i \to j$ if a random walk can move from $i$ to $j$ in finitely many steps. %With these notations we define the essential and inessential states.
\begin{definition}[Essential and inessential state \cite{Levin:09}]
	A state $i \in \Omega$ is essential if for all $j$ such that $i \to j$ it is also true that $j \to i$. A state is inessential if it is not essential.
\end{definition}

Our aim in this section is to establish that if inliers connect to themselves, \ie they are \emph{subspace-preserving} (Section \ref{sec:subspace-preserving}), and the representation $R$ satisfies certain connectivity conditions (Section \ref{sec:connectivity}), then inliers are essential states of the Markov chain and outliers are inessential states. Subsequently, in Section \ref{sec:cesaro_mean_converge} we show that the Ces\`aro mean \eqref{eq:def-I} identifies essential and inessential states, thus establishing the correctness of Algorithm~\ref{alg:main} for outlier detection.

\subsection{Subspace-preserving representation}
\label{sec:subspace-preserving}
We first establish that inliers express themselves with only other inliers when they lie in a union of low dimensional subspaces. This property is well-studied in the subspace clustering literature. We will borrow terminologies and results from prior work and modify them for our current task of outlier detection.
\begin{definition}[Subspace-preserving representation \cite{Vidal:Springer16}]
	If $\x_j \in \S_\ell$ is an inlier, then the representation $\r_j \in \RR^N$ is called subspace-preserving if the nonzero entries of $\r_j$ correspond to points in $\S_\ell$, \ie $r_{ij}\ne0$ only if $\x_i \in \S_\ell$. The representation matrix $R=[\r_1, \cdots, \r_N] \in \RR^{N \times N}$ is called subspace-preserving if $\r_j$ is subspace-preserving for every inlier $\x_j$.
\end{definition}
A representation matrix $R$ is subspace-preserving if each inlier uses points in its own subspace for representation. Given $X$, a subspace-preserving representation $R$ can be obtained by solving \eqref{eq:representation_matrix} when certain geometric conditions hold. The following result is modified from \cite{You:CVPR16-EnSC}. It assumes that columns of $X$ are normalized to have unit $\ell_2$-norm. 
\begin{theorem}\label{thm:subspace-preserving}
	Let $\x_j \in \S_\ell$ be an inlier. Define the oracle point of $\x_j$ to be $\bfdelta_j:=\gamma \cdot (\x_j-X_{-j}^\ell \cdot \r_j^\ell)$, where $X_{-j}^\ell$ is the matrix containing all points in $\S_\ell$ except $\x_j$ and 
	%$\r_j^\ell$ the solution of the following problem
	\begin{equation*}
		\r_j^\ell := \arg\min_{\r} \lambda \|\r\|_1 + \frac{1-\lambda}{2}\|\r\|_2^2 + \frac{\gamma}{2}\|\x_j - X_{-j}^\ell \r\|_2^2.
		%\r_j^\ell := \arg\min_{\r_j^\ell} \lambda \|\r_j^\ell\|_1 + \frac{1-\lambda}{2}\|\r_j^\ell\|_2^2 + \frac{\gamma}{2}\|\x_j - X_{-j}^\ell \r_j^\ell\|_2^2.
		%\label{eq:fictitious}
	\end{equation*}	
The solution $\r_j$ to  \eqref{eq:representation_matrix} is subspace-preserving if
	\begin{equation}\label{eq:condition}
		 \max_{k\ne j,\x_k \in \S_\ell} |\langle \x_k, \bar{\bfdelta}_j\rangle| - \max_{k:\x_k \notin \S_\ell} |\langle \x_k, \bar{\bfdelta}_j\rangle| > \frac{1-\lambda}{\lambda},
	\end{equation}
	where $\bar{\bfdelta}_j := \bfdelta_j / \|\bfdelta_j\|_2$.
\end{theorem}
{An outline of the proof is given in the appendix.} Note that the oracle point $\bfdelta_j$ lies in $\S_\ell$ and that its definition only depends on points in $\S_\ell$. The first term in condition \eqref{eq:condition} captures the distribution of points in $\S_\ell$ near $\bar\bfdelta_j$, and is expected to be large if the neighborhood of $\bar\bfdelta_j$ is well-covered by points from $\S_\ell$. The second term characterizes the similarity between the oracle point $\bar\bfdelta_j$ and all other data points, which includes the outliers and the inliers from other subspaces.  The condition requires the former to be larger than the latter by a margin of $\frac{1-\lambda}{\lambda}$, which is close to zero if $\lambda$ is close to $1$. Overall, condition \eqref{eq:condition} requires that points in $\S_\ell$ are dense around $\bar\bfdelta_j$, which is itself in $\S_\ell$, and that outliers and inliers from other subspaces do not lie close to $\bar\bfdelta_j$.

Even if~\eqref{eq:condition} holds for all $j$ 
%the condition in Theorem \ref{thm:subspace-preserving} is satisfied 
so that the representation $R$ is subspace-preserving, we cannot automatically establish an equivalence between inliers/outliers and essential/inessential states because of potential complications related to the graph's \emph{connectivity}.  This is addressed next.
%in the next section.

\subsection{Connectivity considerations} % assumptions}
\label{sec:connectivity}

 In the context of sparse subspace clustering, the well-known connectivity issue \cite{Nasihatkon:CVPR11,Wang:NIPS13-LRR+SSC,Lu:ICCV13-TraceLasso,You:CVPR16-EnSC,Wang:AISTAT16} refers to the problem that  points in the same subspace may not be well-connected in the representation graph, which may cause oversegmentation of the true clusters.  Thus, one has to make the assumption that each true cluster is connected to guarantee correct clustering. For the outlier detection problem, it may happen that an inlier is inessential and thus classified as an outlier when the inliers are not well-connected; similarly, an outlier may be essential and thus classified as an inlier if it is not connected to at least one inlier.  In fact, the situation is even more involved since the representation graph is directed and inliers and outliers behave differently. 
 %The next section addresses this and formulate connectivity requirements.

Suppose, as a first example, that there exists an inlier that is never used to express any other inliers. This is equivalent to saying that there is no edge going into this point from any other inliers. Note that the subspace-preserving property can still hold if this inlier expresses itself using other inliers.  Yet, since a random walk leaving this point would never return it can not be identified as an inlier. To avoid such cases, we need the following assumption.
\begin{assumption}
	%For any two points in the same subspace, the corresponding vertices in the representation graph is strongly connected,  i.e. there is a path in each direction between them.
	For any inlier subspace $\S_\ell$, the vertices $\{\x_j \in \S_\ell\}$ of the representation graph are strongly connected, i.e. there is a path in each direction between each pair of vertices.
	\label{asmp:inlier}
\end{assumption}

%Assumption \ref{asmp:inlier} requires good connectivity between points from the same inlier subspace. As an extreme example, it is trivially satisfied if points from each subspace are densely connected, i.e, for all $j$ such that  $\x_j$ is an inlier, $r_{ji}\ne 0$ for all $i$ such that $j\ne i$ and $\x_i$ is in the same subspace as $\x_j$. From this perspective, it is reasonable to expect that if the graph is denser then the assumption is more likely to be satisfied. In reality, there is often a trade-off between getting densely connected graph and subspace-preserving representations \cite{Wang:NIPS13-LRR+SSC,You:CVPR16-EnSC}, \ie as one constructs more connections, it becomes more likely that some connections break subspace-preserving property. In this work, we choose the Elastic net for constructing representation \eqref{eq:representation_matrix} for which the trade-off effect has been studied in \cite{You:CVPR16-EnSC}.

Assumption \ref{asmp:inlier} requires good connectivity between points from the same inlier subspace.
We also %In addition to Assumption~\ref{asmp:inlier}, we also 
need good connectivity between outliers and inliers. Consider the example when there is a subset of outliers for which all of their outgoing edges lead only to points within that same subset.
%, i.e., each one of them is expressed by others that are also in this subset.
In this case, the subset of points can not be detected as outliers since their representation pattern is the same as for the inliers. The next assumption rules out this case.
\begin{assumption}
	For each subset of outliers there exists an edge in the representation graph that goes from a point in this subset to an inlier or to an outlier outside this subset.
	\label{asmp:outlier}
\end{assumption}

\subsection{Main theorem: guaranteed outlier detection}
\label{sec:cesaro_mean_converge}

We can now establish guaranteed outlier detection by our representation graph based method stated as Algorithm~\ref{alg:main}.
\begin{theorem}\label{thm:main}
	If the representation $R$ is subspace-preserving and satisfies Assumptions \ref{asmp:inlier} and \ref{asmp:outlier}, then Algorithm~\ref{alg:main} with $T=\infty$ and $\epsilon=0$ correctly identifies outliers.
\end{theorem}
Theorem~\ref{thm:main} is a direct consequence of the following two facts whose proofs are provided in {the appendix}. \nocite{Gallager:13,Tijms:03}
\begin{lemma}\label{thm:inlier-essential-equivalence}
	If the representation $R$ is subspace-preserving and Assumptions \ref{asmp:inlier} and \ref{asmp:outlier} hold, then inliers and outliers correspond to essential and inessential states, respectively.
	\end{lemma}
\begin{lemma}\label{thm:cesaro-mean-converge}
	For any probability transition matrix $P$, the averaged probability distribution in \eqref{eq:def-I} satisfies $\lim_{T\to \infty} \bar{\bfpi}^{(T)} = \bfpi$, where $\bfpi$ is such that $\pi_j = 0$ if and only if state $j$ is inessential.
\end{lemma}

Theorem~\ref{thm:main} shows that Problem~\ref{pr:outlier-detection} is solved by Algorithm~\ref{alg:main} if the data $X$ satisfies the geometric conditions in \eqref{eq:condition} and the representation graph satisfies the required connectivity assumptions. 

We note that the random walk by the Ces\`aro mean adopted here is different from the popular random walk with restart as adopted by PageRank, for example. The benefit of PageRank is that the random walk converges to the unique stationary distribution. However, it is not clear whether this stationary distribution identifies the outliers. In fact, all states in the random walk of PageRank are essential, so that outliers do not converge to zero probabilities. In contrast, the random walk in our method does not necessarily have a unique stationary distribution, but the Ces\`aro mean does converge to one of the stationary distributions, which we have shown can be used to identify outliers. {A detailed discussion is in the Appendix.}

%\begin{lemma}
%	If the representation $R$ is normal, then inliers correspond to essential states and outliers correspond to inessential states, \ie $\x_j$ is inlier if and only if 
%\end{lemma}
%
%\begin{lemma}[\cite{Gallager:13}]
%	The probability transition matrix $P$ satisfies $\lim_{r \to \infty} I(r) = \pi$, where $\pi$ is such that $\pi_j = 0$ if and only if state $j$ is inessential.
%\end{lemma}

\section{Experiments}
\label{sec:experiment}

\begin{figure}[t]
	\centering
	\subfigure[Extended Yale B]{\includegraphics[scale = 0.28,trim={0.5cm 8cm 3cm 0cm},clip]{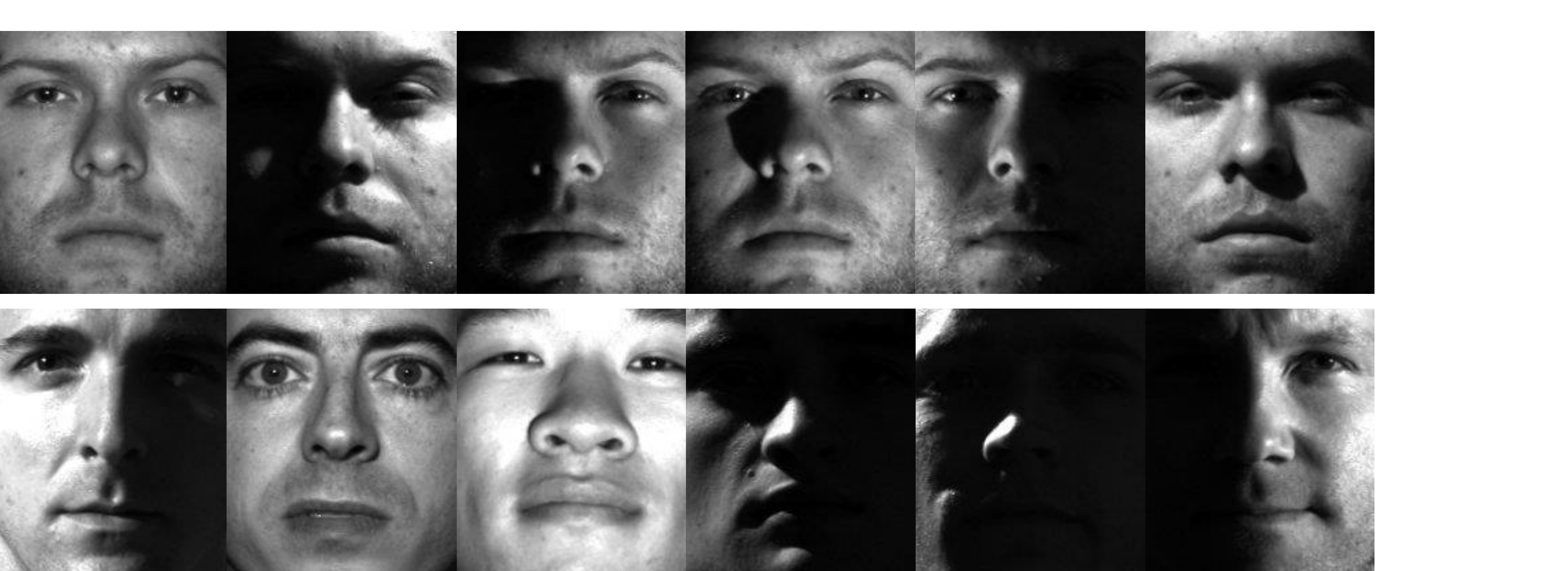}}\\
	\subfigure[Caltech-256]{\includegraphics[scale = 0.31,trim={0cm 11cm 6cm 0.5cm},clip]{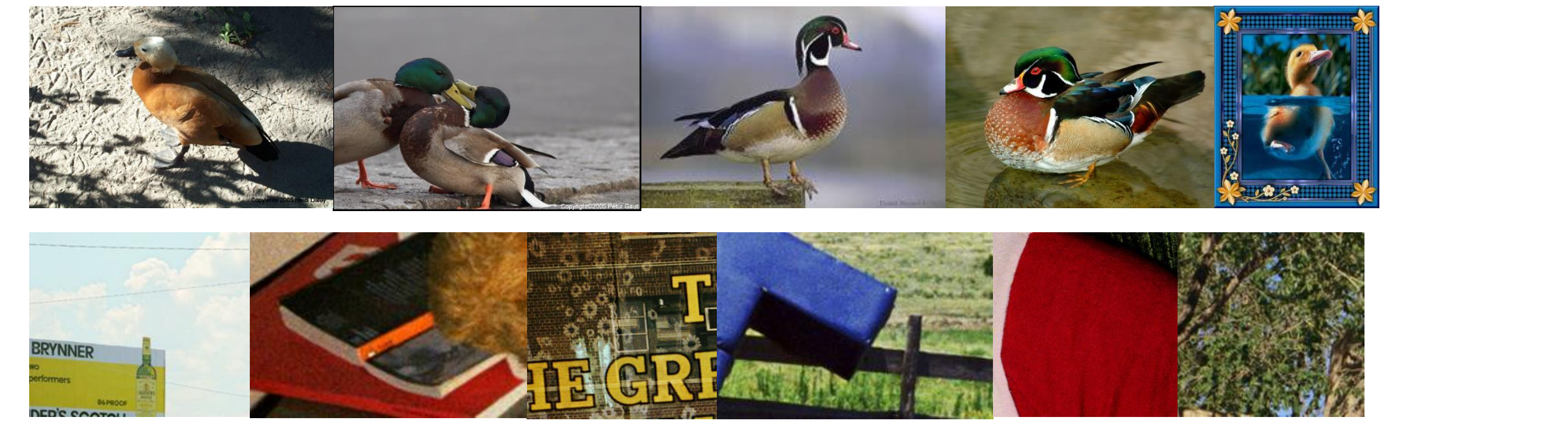}}\\
	\subfigure[Coil-100]{\includegraphics[scale = 0.36,trim={0.5cm 11cm 8cm 1cm},clip]{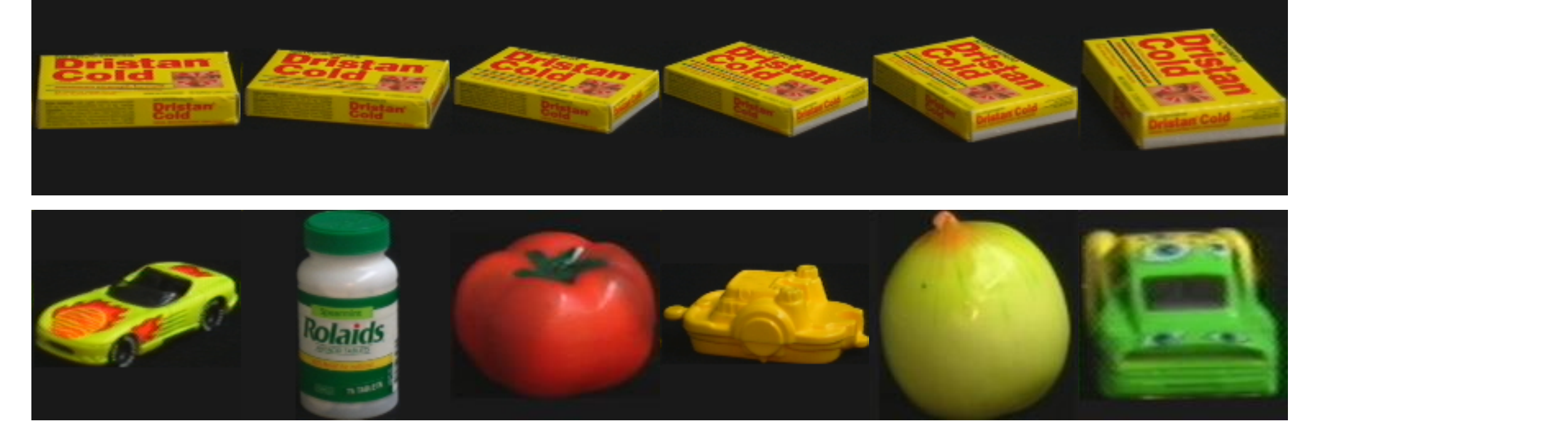}}
	\caption{Examples of data used for outlier detection. For each database, the top row shows examples of the inlier set and the bottom row shows examples from the outlier set. }
	\label{fig:data}
%	\vspace{-2mm}
\end{figure}

We  use several image databases (see Figure~\ref{fig:data}) to evaluate our outlier detection method (Algorithm~\ref{alg:main}). For computing the representation $\r_j$ in \eqref{eq:representation_matrix}, we use the solver in \cite{Jin:IP09} with $\lambda=0.95$ and $\gamma= \alpha \cdot \frac{\lambda}{\max_{i:i\ne j}|\x_j^\transpose \x_i|}$, where $\alpha$ is a parameter tuned to each dataset. In particular, the solution to \eqref{eq:representation_matrix} is nonzero if and only if $\alpha>1$. The number of iterations $T$ is set to be $1,\!000$.

\begin{table*}[t]
	\centering
	\caption{Results on the Extended Yale B database. Inliers are taken to be the images of either one or three randomly chosen subjects, and outliers are randomly chosen from the other subjects (at most one from each subject). For R-graph we set $\alpha=5$ in the definition of $\gamma$. }
	\label{tbl:result_EYaleB}
	\begin{tabular}{c||c|c|c|c|c|c|c|c}
		    & OutRank &  CoP  & REAPER & OutlierPursuit &  LRR  & DPCP  & $\ell_1$-thresholding &           R-graph (ours)            \\ \hline
		\multicolumn{9}{l}{\textsl{Inliers: all images from \textbf{one} subject ~~ Outliers: $35\%$, taken from other subjects }}    \\ \hline
		AUC &  0.536  & 0.556 & 0.964  &     0.972      & 0.857 & 0.952 &         0.844         &           \textbf{0.986}            \\ \hline
		F1  &  0.552  & 0.563 & 0.911  &     0.918      & 0.797 & 0.885 &         0.763         &           \textbf{0.951}            \\ \hline
		\multicolumn{9}{l}{\textsl{Inliers: all images from \textbf{three} subjects ~~ Outliers: $15\%$, taken from other subjects }} \\ \hline
		AUC &  0.519  & 0.529 & 0.932  &     0.968      & 0.807 & 0.888 &         0.848         &           \textbf{0.985}            \\ \hline
		F1  &  0.288  & 0.292 & 0.758  & 0.856 & 0.509 & 0.653 &         0.545         &                \textbf{0.878}                \\ \hline
	\end{tabular}
\end{table*}

\subsection{Experimental setup} 

\myparagraph{Databases} We construct outlier detection tasks from three publicly available databases. The Extended Yale B \cite{Georghiades2001-PAMI} dataset contains frontal face images of $38$ individuals each under $64$ different illumination conditions. The face images are of size $192 \times 168$, for which we downsample to $48 \times 42$. The Caltech-256  \cite{Griffin:Caltech256} is a database that contains images from $256$ categories that have more than $80$ images each. There is also an additional ``clutter'' category in this database that contains $827$ images of different varieties, which are used as outliers. The Coil-100 dataset \cite{Nene:1996-coil} contains $7,\!200$ images of $100$ different objects. Each object has $72$ images taken at pose intervals of $5$ degrees, with the images being of size $32 \times 32$. For the Extended Yale B and Coil-100 datasets we use raw pixel intensity as the feature representation. Images in Caltech-256 are represented by a $4,\!096$-dimensional feature vector extracted from the last fully connected layer of the 16-layer VGG network \cite{Simonyan:Arxiv14}.

\myparagraph{Baselines} We compare with $6$ other representative methods that are designed for detecting outliers in one or multiple subspaces: CoP \cite{Rahmani:arXiv16}, OutlierPursuit \cite{Xu:NIPS10}, REAPER \cite{Lerman:FCM15}, DPCP \cite{Tsakiris:DPCPICCV15}, LRR \cite{Liu:ICML10} and $\ell_1$-thresholding \cite{Soltanolkotabi:AS12}. We also compare with a graph based method: OutRank \cite{Moonesinghe:ICTAI06,Moonesinghe:IJAIT08}. We implement the inexact ALM \cite{Lin:09} for solving the optimization in OutlierPursuit. For LRR, we use the code available online at \url{https://sites.google.com/site/guangcanliu/}. For DPCP, we use the code provided by the authors. All other methods are implemented according to the description in their respective papers. 

%We also compare with OutRank \cite{Moonesinghe:ICTAI06,Moonesinghe:IJAIT08}, which is also a random walk method for outlier detection. It is hypothesized in OutRank that outliers are points that have low similarity to other data points. To find such points, a graph is constructed in which the weight of an edge is set to the cosine similarity or RBF similarity between the two connected data points. Then, a random walk algorithm is used and outliers are those that end up with low probability. Note that OutRank is presented as a general outlier detection method \emph{without} the assumption that inliers come from subspaces.

\myparagraph{Evaluation metric} Each outlier detection method generates a numerical value for each data point that indicates its ``outlierness'', and a threshold value is required for determining inliers and outliers. A Receiver Operating Characteristic (ROC) curve plots the true positive rate and false positive rate for  all threshold values. We use the area under the curve (AUC) as a metric of performance in terms of the ROC. The AUC is always between $0$ and $1$, with a perfect model having an AUC of $1$ and 
%one if there exists a threshold that can perfectly separate inliers and outliers. For  
a model that guesses randomly having an AUC of approximately $0.5$.

As a second metric, we provide the F1-score, which is the harmonic mean of precision and recall. The F1-score is dependent upon the threshold, and we report the largest F1-score across all thresholds. An F1-score of $1$ means there exists a threshold that gives both precision and recall equal to $1$, \ie a perfect separation of inliers and outliers.

The reported numbers for all experiments discussed in this section are the averages over 50 trials.

\subsection{Outliers in face images}

\begin{table*}[t]
	\centering
	\caption{Results on the Caltech-256 database. Inliers are taken to be images of one, three, or five randomly chosen categories, and outliers are randomly chosen from category 257-clutter. For R-graph we set $\alpha=20$ in the definition of $\gamma$. }
	\label{tbl:result_Caltech256}
	\begin{tabular}{c||c|c|c|c|c|c|c|c}
		& OutRank &  CoP  & REAPER & OutlierPursuit &  LRR  & DPCP  & $\ell_1$-thresholding &  R-graph (ours)  \\ \hline
		\multicolumn{9}{l}{\textsl{Inliers: \textbf{one} category of images ~~ Outliers: $50\%$ }}     \\ \hline
		AUC &  0.897  & 0.905 & 0.816  &     0.837      & 0.907 & 0.783 &  0.772   &  \textbf{0.948}  \\ \hline
		F1  &  0.866  & 0.880 & 0.808  &     0.823      & 0.893 & 0.785 &  0.772   &  \textbf{0.914}  \\ \hline
		\multicolumn{9}{l}{\textsl{Inliers: \textbf{three} categories of images ~~ Outliers: $50\%$ }} \\ \hline
		AUC &  0.574  & 0.676 & 0.796  &     0.788      & 0.479 & 0.798 &  0.810   &  \textbf{0.929}  \\ \hline
		F1  &  0.682  & 0.718 & 0.784  &     0.779      & 0.671 & 0.777 &  0.782   &  \textbf{0.880}  \\ \hline
		\multicolumn{9}{l}{\textsl{Inliers: \textbf{five} categories of images ~~ Outliers: $50\%$ }}  \\ \hline
		AUC &  0.407  & 0.487 & 0.657  &     0.629      & 0.337 & 0.676 &  0.774   &  \textbf{0.913}  \\ \hline
		F1  &  0.667  & 0.672 & 0.716  &     0.711      & 0.667 & 0.715 &  0.762   &  \textbf{0.858}  \\ \hline
	\end{tabular}
\end{table*}
\begin{table*}[t]
	\centering
	\caption{Results on the Coil-100 database. Inliers are taken to be images of one, four, or seven randomly chosen categories, and outliers are randomly chosen from other categories (at most one from each category). For R-graph we set $\alpha=10$ in the definition of $\gamma$.}
	\label{tbl:result_Coil100}
	\begin{tabular}{c||c|c|c|c|c|c|c|c}
		    & OutRank &  CoP  & REAPER & OutlierPursuit &  LRR  & DPCP  & $\ell_1$-thresholding & R-graph (ours) \\ \hline
		\multicolumn{9}{l}{\textsl{Inliers: all images from \textbf{one} category ~~ Outliers: $50\%$ }}          \\ \hline
		AUC &  0.836  & 0.843 & 0.900  &     0.908      & 0.847 & 0.900 &         0.991         & \textbf{0.997} \\ \hline
		F1  &  0.862  & 0.866 & 0.892  &     0.902      & 0.872 & 0.882 &         0.978         & \textbf{0.990} \\ \hline
		\multicolumn{9}{l}{\textsl{Inliers: all images from \textbf{four} categories ~~ Outliers: $25\%$ }}       \\ \hline
		AUC &  0.613  & 0.628 & 0.877  &     0.837      & 0.687 & 0.859 &         0.992         & \textbf{0.996} \\ \hline
		F1  &  0.491  & 0.500 & 0.703  &     0.686      & 0.541 & 0.684 &         0.941         & \textbf{0.970} \\ \hline
		\multicolumn{9}{l}{\textsl{Inliers: all images from \textbf{seven} categories ~~ Outliers: $15\%$ }}      \\ \hline
		AUC &  0.570  & 0.580 & 0.824  &     0.822      & 0.628 & 0.804 &         0.991         & \textbf{0.996} \\ \hline
		F1  &  0.342  & 0.346 & 0.541  &     0.528      & 0.366 & 0.511 &         0.897         & \textbf{0.955} \\ \hline
	\end{tabular}
\end{table*}

Suppose we are given a set of images of one or more individuals but that the data set is also corrupted by face images of a variety of other individuals. The task is to detect and remove those outlying face images. It is known that images of a face under different lighting conditions lie approximately in a low dimensional subspace. Thus, this task can be modeled as the problem of outlier detection in one subspace or in a union of subspaces.

We use the extended Yale B database. In the first experiment, we randomly choose a single individual from the 38 subjects and use all 64 images of this subject as the inliers. We then choose images from the remaining 37 subjects as outliers with at most one image from each subject. The overall data set has $25\%$ outliers. The average AUC and F1 measures over 50 trials are reported in Table~\ref{tbl:result_EYaleB}. For a fair comparison, we fine-tuned the parameters for all methods.

\myparagraph{Comparing to state of the art} We see that our representation graph based method R-graph outperforms the other methods.
%, demonstrating the effectiveness of our method. 
Besides our method, the REAPER, Outlier Pursuit and DPCP algorithms all perform well. These three methods learn a single subspace and treat those that do not fit the subspace as outliers, thus making them well suited for this data (the images of one individual can be well-approximated by a single low dimensional subspace). 

%The LRR can be viewed as a variation of Outlier Pursuit, with the main differences being  that LRR uses self-expressiveness to model subspaces and uses a low rank regularizer on the coefficient matrix. By exploiting self-expressiveness, LRR has the benefit that it is able to deal with multiple subspaces. However, our experiments show that its performance is not as good as Outlier Pursuit. This can probably be explained by the fact that the algorithm for solving the LRR model has no convergence guarantee. %, so the solution may not be optimal. 

%The $\ell_1$-thresholding does not give very good results even though it also uses sparse representation as in our method. This shows that the magnitude of the representation vector is not a robust measure for classifying outliers, and by considering the connecting patterns in the representation graph, our method can significantly improve the results.

The LRR and $\ell_1$-thresholding methods use data self-representation, which is also the case for our method.  However, LRR does not give good outlier detection results, probably because its algorithm for solving the LRR model is not guaranteed to converge to a global optimum. The $\ell_1$-thresholding also does not give good results, showing that the magnitude of the representation vector is not a robust measure for classifying outliers. By considering the connection patterns in the representation graph, our method achieves significantly better results.

The performance of OutRank and CoP is significantly worse than that of the other methods.  This poor performance can be explained by the use of a coherence-based distance, which fails to capture similarity between data points when the data lie in subspaces. For example, it can be argued that the coherence between two faces with the same illumination condition can be higher than two images of the same face under different illumination conditions. 

\myparagraph{Dealing with multiple inlier groups} In order to test the ability of the methods to deal with multiple inlier groups, we designed a second experiment in which inliers are taken to be images of $3$ randomly chosen subjects, and outliers are randomly drawn from other subjects as before. For all methods, we use the same parameters as in the previous experiment to test the robustness to parameter tuning. The results of this experiment are reported in Table~\ref{tbl:result_EYaleB}. 

We can see that Outlier Pursuit and our R-graph are the two best methods. Although Outlier Pursuit only models a single low dimensional subspace, it can still deal with this data since the union of the three subspaces corresponding to the three subjects in the inlier set is still low dimensional and can be treated as a single low dimensional subspace. However, we postulate that Outlier Pursuit will eventually fail as we increase the number of inlier groups, since the union of low dimensional subspaces will no longer be low rank. Our method does not have this limitation.

% and is able to deal with cases where the union of inlier subspaces is no longer low rank.

Similar to Outlier Pursuit, both REAPER and DPCP can, in principle, handle multiple inlier groups by fitting a single subspace to their union. However,  REAPER and DPCP require as input the dimension of the union of the inlier subspaces, which can be hard to estimate in practice. Indeed, in Table~\ref{tbl:result_EYaleB}, we observe that the performances of REAPER and DPCP are less competitive in comparison to Outlier Pursuit and our R-graph for the three subspace case.

%\myparagraph{Choosing the parameter}
%[how to choose parameters, robustness to parameters]

\subsection{Outliers in images of objects}

We test the ability of the methods to identify one or several object categories that frequently appear in a set of images amidst outliers that consist of objects that rarely occur.  For Caltech-256, images in $n \in \{1, 3, 5\}$ randomly chosen categories are used as inliers in three different experiments. From each category, we use the first $150$ images if the category has more than $150$ images. We then randomly pick a certain number of images from the ``clutter'' category as outliers such that there are $50\%$ outliers in each experiment. For Coil-100, we randomly pick $n \in \{1, 4, 7\}$ categories as inliers and pick at most one image from each of the remaining categories as outliers. 

The results are reported in Table~\ref{tbl:result_Caltech256} and Table \ref{tbl:result_Coil100}.  We see that our R-graph method achieves the best performance. The two geometric distance based methods, OutRank and CoP, achieve good results when there is one inlier category, but deteriorate when the number of inlier categories increases. The performance of REAPER, Outlier Pursuit and DPCP are similar to each other and worse than our method. This may be because they all try to fit a linear subspace to the data, while the data in these two databases may be better modeled by a nonlinear manifold. The $\ell_1$-thresholding and the representation graph method are all based on data self-expression, and seem to be more powerful for this data.

\section{Conclusion}

We presented an outlier detection method that combined data self-representation and random walks on a representation graph. Unlike many prior methods for robust PCA, our method is able to deal with multiple subspaces and does not require the number of subspaces or their dimensions to be known. Our analysis showed that the method is guaranteed to identify outliers when certain geometric conditions are satisfied and two connectivity assumptions hold. In our experiments on face image and object image databases, our method achieves the state-of-the-art performance.  

\section*{Acknowledgment}

This work was supported by NSF BIGDATA grant 1447822. The authors also thank Manolis Tsakiris, Conner Lane and Chun-Guang Li for helpful comments.

\newpage

\numberwithin{equation}{section}
\numberwithin{figure}{section}
\numberwithin{table}{section}

\begin{appendices}
	
The appendix is organized as follows. In Section \ref{sec:prf-subspace-preserving}  we discuss subspace-preserving representations and give an outline of the proof for Theorem~\ref{thm:subspace-preserving}. Section \ref{sec:markov-chain} contains relevant background on Markov chain theory, which is then used in Section \ref{sec:prf-main} for proving Lemma~\ref{thm:inlier-essential-equivalence} and Lemma~\ref{thm:cesaro-mean-converge}, as well as providing an in-depth discussion of the Ces\`aro mean used for outlier detection. In Section \ref{sec:additional-experiment} we provide some additional results for experiments on the Extended Yale B database that provide additional insight into the behavior of the methods.
	
\section{Subspace-preserving representation and proof of Theorem~\ref{thm:subspace-preserving}}
\label{sec:prf-subspace-preserving}
	
The idea of a subspace-preserving representation has been extensively studied in the literature of subspace clustering to guarantee the correctness of clustering \cite{Elhamifar:TPAMI13,Soltanolkotabi:AS12,Soltanolkotabi:AS14,Lu:ECCV12,Liu:PAMI12,Dyer:JMLR13,Park:NIPS14,You:ICML15,Heckel:TIT15,You:CVPR16-SSCOMP,You:CVPR16-EnSC,Wang:ICML15,Wang:JMLR16,Li:AAAI17}. Concretely, the data in a subspace clustering task are assumed to lie in a union of low dimensional subspaces, without any outliers that lie outside of the subspaces. A data self-representation matrix is called subspace-preserving if each point uses only points that are from its own subspace in its representation. 
	
Theoretical results in subspace clustering can be adapted to study subspace-preserving representations in the presence of outliers. Here, we use the analysis and result from \cite{You:CVPR16-EnSC}, which studied the elastic net representation \eqref{eq:representation_matrix} for subspace clustering, to prove a subspace-preserving representation result in the presence of outliers, \ie Theorem~\ref{thm:subspace-preserving}. We also present a corollary of Theorem~\ref{thm:subspace-preserving} which allows us to compare our result with other subspace clustering results.
	
\subsection{Proof of Theorem~\ref{thm:subspace-preserving}}
	
The proof of Theorem~\ref{thm:subspace-preserving} follows mostly from the work \cite{You:CVPR16-EnSC}. We provide an outline of the proof for completeness.

Consider the vector $\r_j^\ell$, which is the solution of the problem in the statement of Theorem~\ref{thm:subspace-preserving}. Notice that the entries of $\r_j^\ell$ correspond to columns of the data matrix $X_{-j}^\ell$. One can subsequently construct a representation vector by padding additional zeros to $\r_j^\ell$ at entries corresponding to points in $X$ that are not in $X_{-j}^\ell$. Note that this vector is trivially subspace-preserving by construction. The idea of the proof is to show that this constructed vector, which is subspace-preserving by construction, is a solution to the optimization problem \eqref{eq:representation_matrix} (and no other vector is).
A sufficient condition for this to hold is that $\bfdelta_j$, which is computed from $\r_j^\ell$, needs to have low correlation with all points $\x_k \notin \S_\ell$. More precisely, we have the following lemma.

\begin{lemma}[{\cite[Lemma 3.1]{You:CVPR16-EnSC}}]\label{thm:subspace-preserving-lemma}
	The vector $\r_j$ is subspace-preserving if $|\langle \x_k, \bfdelta_j \rangle| < \lambda$ for all $\x_k \notin \S_\ell$.
\end{lemma}
Lemma \ref{thm:subspace-preserving-lemma} can be proved by using the optimality condition of the optimization problem in \eqref{eq:representation_matrix}. Equivalently, it suggests that $\r_j$ is subspace-preserving if 
\begin{equation}
\max_{k: \x_k \notin \S_\ell}|\langle \x_k, \bar{\bfdelta}_j \rangle| < \frac{\lambda}{\|\bfdelta_j\|_2}.
\label{eq:subspace-preserving-condition}
\end{equation}
To get more meaningful results, we need an upper bound on $\|\bfdelta_j\|_2$. This is provided by the following lemma.
\begin{lemma}[{\cite[Lemma C.2]{You:arxiv16-EnSC}}]
	If $\kappa_j$ be the maximum coherence between the oracle point $\bfdelta_j$ and columns of $X_{-j}^\ell$, i.e. $\kappa_j=\max_{k\ne j: \x_k \in \S_\ell}|\langle \x_k, \bar{\bfdelta}_j\rangle|$, then
	\begin{equation}
	\|\bfdelta_j\|_2 \le \frac{\lambda\kappa_j + 1-\lambda}{\kappa_j^2}.
	\label{eq:bound-delta}
	\end{equation}
\end{lemma}

Combining \eqref{eq:subspace-preserving-condition} and \eqref{eq:bound-delta}, $\r_j$ is subspace-preserving if
\begin{equation}
\max_{k: \x_k \notin \S_\ell}|\langle \x_k, \bar{\bfdelta}_j \rangle| <  \frac{\kappa_j^2}{\kappa_j + \frac{1-\lambda}{\lambda}}.
\end{equation}
To simplify the result, note that
%\begin{multline}
\begin{align*}
\frac{\kappa_j^2}{\kappa_j + \frac{1-\lambda}{\lambda}} 
&= \kappa_j \cdot \left(\frac{1}{1 + \frac{1-\lambda}{\lambda}\frac{1}{\kappa_j}}\right) \\
&\ge \kappa_j \cdot\left(1 - \frac{1-\lambda}{\lambda}\frac{1}{\kappa_j}\right) = \kappa_j - \frac{1-\lambda}{\lambda}.
\end{align*}
%\end{multline}
Therefore, a sufficient condition for $\r_j$ to be subspace-preserving is that
\begin{equation}
\max_{k: \x_k \notin \S_\ell}|\langle \x_k, \bar{\bfdelta}_j \rangle| <\kappa_j - \frac{1-\lambda}{\lambda}.
\label{eq:condition-proof}
\end{equation}
Since \eqref{eq:condition-proof} is the same as \eqref{eq:condition}, the proof has been completed.
%. This finishes the proof.

\subsection{Discussions}

Another commonly used geometric quantity for characterizing when representations will be subspace-preserving is the inradius of sets of points \cite{Soltanolkotabi:AS12,Soltanolkotabi:AS14,You:ICML15,You:CVPR16-SSCOMP,Wang:NIPS13-LRR+SSC,Wang:JMLR16,Wang:ICML15}. In order to understand the relationship to the results found in these works, we present a corollary of Theorem~\ref{thm:subspace-preserving}.
\begin{definition}[inradius]
	The (relative) inradius of a convex body $\P$, denoted as $\rho(\P)$, is the radius of the largest $\ell_2$ ball in the span of $\P$ that can be inscribed in $\P$.
\end{definition}

\begin{corollary}\label{thm:subspace-preserving-inradius}
	If $\x_j \in \S_\ell$ is an inlier, then $\r_j$ computed from \eqref{eq:representation_matrix} is subspace-preserving if
	\begin{equation}
	\rho_j - \max_{k:\x_k \notin \S_\ell} |\langle \x_k, \bar{\bfdelta}_j\rangle| > \frac{1-\lambda}{\lambda},
	\label{eq:condition-inradius}
	\end{equation}
	where $\bar{\bfdelta}_j$ is defined in  Theorem~\ref{thm:subspace-preserving}, and $\rho_j$ is the inradius of the convex hull of  the symmetrized points in $X_{\-j}^\ell$, \ie
	\begin{equation}
	\rho_j := \rho(\conv\{\pm \x_k: \x_k \in \S_\ell, k \ne j\}).
	\end{equation}
\end{corollary}
The inradius captures the distribution of the columns of $X_{-j}^\ell$, \ie it is large if points are well spread out in $\S_\ell$. Thus, the condition in \eqref{eq:condition-inradius} is easier to be satisfied if the set of points in $\S_\ell$ is dense and well covers the entire subspace. Note that this requirement is stronger than that in Theorem~\ref{thm:subspace-preserving}, which only requires points in $\S_\ell$ to be dense around the oracle point $\delta_j$ (\ie it requires $\max_{k\ne j: \x_k \in \S_\ell}|\langle \x_k, \bar{\bfdelta}_j\rangle| $ to be large). In fact, it is established in \cite{You:CVPR16-EnSC} that $\max_{k\ne j: \x_k \in \S_\ell}|\langle \x_k, \bar{\bfdelta}_j\rangle| \ge \rho_j$, so that the condition in \eqref{eq:condition-inradius}  is a stronger requirement than that of \eqref{eq:condition} in Theorem~\ref{thm:subspace-preserving}.

\section{Background on Markov chain theory}
\label{sec:markov-chain}

We present background material on Markov chain theory that will help us understand the Ces\`aro mean \eqref{eq:def-I} used for outlier detection in our method. The following material is organized from textbooks \cite{Serfozo:09,Gallager:13,Tijms:03,Levin:09} and the website \url{http://www.math.uah.edu/stat}.

We consider a Markov chain $(X_0, X_1, \cdots)$ on a finite state space $\Omega$ with transition probabilities $p_{ij}$ for $i,j \in \Omega$. The $t$-step transition probabilities are defined to be $p_{ij}^{(t)}:= P\{X_t=j | X_0=i\}$.

\subsection{Decomposition of the state space}

%\myparagraph{Decomposition of the state space} 
A Markov chain can be decomposed into more basic and manageable parts.

\begin{definition}
	State  $j$ is accessible from state $i$, denoted as $i \to j$, if $p_{ij}^{(t)}>0$ for some $t> 0$.  We say that the states $i$ and $j$ communicate with each other, denoted by $i \leftrightarrow j$, if $i \to j$ and $j \to i$.
\end{definition}

Since it can be shown that $\leftrightarrow$ is an equivalence relation, it induces a partition of the state space $\Omega$ into disjoint equivalence classes known as \emph{communicating classes}. We are interested in each of the \emph{closed} communicating classes.
\begin{definition}
	A non-empty set $C \subseteq \Omega$ is called a closed set if $p_{ij}=0$ for $i \in C$ and $j \notin C$.
\end{definition}
Note that states in a closed communicating class are essential while states in other communicating classes are inessential \cite{Levin:09}.
\begin{theorem}[\cite{Serfozo:09}]\label{thm:decomposition}
	The state space $\Omega$ has the unique decomposition $\Omega = \I \cup \E_1 \cup \dots \E_n$, where $\I$ is the set of inessential states, and $\E_1, \dots, \E_n$ are closed communicating classes containing essential states.
\end{theorem}
By Theorem~\ref{thm:decomposition}, the state space of any Markov chain is composed of the essential states and inessential states, and the essential states can be further decomposed into a union of communicating classes. Therefore, the probability transition matrix $P$ can be written in the following form (up to permutation of the states):
\begin{equation}
P = \left[ \begin{array}{cccc}
P_{\E_1\to\E_1} &        & \0 & \0\\
& \ddots &    & \vdots\\
\0  &        & P_{\E_n\to\E_n}& \0\\
P_{\I\to\E_1}&\cdots& P_{\I\to\E_n}&P_{\I\to\I}\\
\end{array} \right]
\label{eq:decomposition-P}
\end{equation}

\subsection{Stationary distribution}

%\myparagraph{Stationary distribution} 
A nonnegative row vector $\bfpi$ is called a \emph{stationary distribution} for the Markov chain if it satisfies $\bfpi = \bfpi P$.
\begin{theorem}[{\cite[Proposition 1.14, Corollary 1.17]{Levin:09}}]\label{thm:stationary-distribution}
	A Markov chain consisting of one closed communicating class has a unique stationary distribution. Moreover, each entry of the stationary distribution is positive.
\end{theorem}
By Theorem~\ref{thm:stationary-distribution}, each component $\E_\ell$ for $\ell=1, \cdots, n$ in the decomposition of the Markov chain in Theorem~\ref{thm:decomposition} has a unique positive stationary distribution $\bfpi_{\E_\ell}$, \ie
\begin{equation}\label{eq:stationary-distribution}
\bfpi_{\E_\ell} = \bfpi_{\E_\ell} \cdot P_{\E_\ell \to \E_\ell}
\ \ \text{with} \ \  \bfpi_{\E_\ell} > 0 \ \text{and} \  \sum_j (\bfpi_{\E_\ell})_j = 1.
%\bfpi_{\E_\ell} ~\text{being positive}.
\end{equation}
%with $\bfpi_{\E_\ell} > 0$ and $\sum_j (\bfpi_{\E_\ell})_j = 1$.
We may then define a stationary distribution for $P$ as 
\begin{equation}\label{eq:stationary-distribution-decomposition}
[\alpha_1 \bfpi_{\E_1}, \dots, \alpha_n \bfpi_{\E_n}, \0] 
\ \ \text{for any} \ \ \alpha_\ell \ge 0, \ \sum_{\ell=1}^{n} \alpha_\ell = 1.
\end{equation}
Note that there is not a unique stationary distribution for $P$ when $n \ge 2$.

\subsection{Convergence of the Ces\`aro mean $\frac{1}{T} \sum_{t=1} ^T P^t$}
%\myparagraph{Convergence of Ces\`aro mean} 
%We present results on the convergence of the Ces\`aro mean $\frac{1}{T} \sum_{t=1} ^T P^t$.

Let $f_{ij}^{(t)}:=P\{ X_t = j, X_{t'} \ne j \text{ for } 1 \le t' < t | X_0 = i\}$ be the probability that the chain starting at $i$ enters $j$ for the first time at the $t$-th step.
The \emph{hitting probability} $f_{ij}=P\{X_t = j \text{ for some } t > 0 | X_0 = i\}$ is the probability that the random walk ever makes a transition to state $j$ when started at $i$, \ie
\begin{equation}
f_{ij} = \sum_{t=1}^\infty f_{ij}^{(t)}.
\end{equation}

The \emph{mean return time} $\mu_{j}:= \sum_{t=1}^{\infty} t f_{jj}^{(t)}$ is the expected time for a random walk starting from state $j$ will return to state $j$.
A general convergence result is stated as follows.
\begin{theorem}[{\cite[Theorem 3.3.1]{Tijms:03}}]	\label{thm:cesaro-convergence}
	For any $i,j \in \Omega$,
	\begin{equation}
	\lim_{T \to \infty} \frac{1}{T} \sum_{t=1}^{T} p_{ij}^{(t)} = \frac{f_{ij}}{\mu_{j}}.
	\end{equation}
\end{theorem}
This result can be simplified by using the decomposition in Theorem~\ref{thm:decomposition}, which leads to the following lemma.
\begin{lemma}	\label{thm:hitting-time}
	If $i,j\in \Omega$ are in the same closed communicating class, then $f_{ij}=f_{ji}=1$.  Also, if $i \in \Omega$ is an inessential state and $\E_\ell \subseteq \Omega$ is a closed communicating class, then $f_{ij} = f_{i \to \E_\ell}$ for all $j \in \E_\ell$, where
	$f_{i \to \E_\ell}$ is the hitting probability from state $i$ to class $\E_\ell$.
\end{lemma}

The following result relates the mean return time with the stationary distribution.
\begin{lemma}	\label{thm:mean-return-time}
	For every closed communicating class $\E_\ell \subseteq \Omega$, it holds that $\bfmu_{\E_\ell} = 1 /  \bfpi_{\E_\ell}$ (entry-wise division), where $\bfmu_{\E_\ell}$ is the vector of mean return times of states in $\E_\ell$.
	If $i \in \Omega$ is an inessential state, then $\mu_{i}=\infty$.
\end{lemma}

By combining Theorem~\ref{thm:cesaro-convergence} with Lemma~\ref{thm:hitting-time} and Lemma~\ref{thm:mean-return-time}, the Ces\`aro limit of a probability transition matrix of the form in \eqref{eq:decomposition-P} can be written as
\begin{multline}
\lim_T \frac{ 1}{T}\sum_{t=1}^T P^t  = \\\left[ \begin{array}{cccc}
\1\cdot \bfpi_{\E_1} &        & \0 & \0\\
& \ddots &    & \vdots\\
\0  &        & \1\cdot \bfpi_{\E_n}& \0\\
\bff_{\I \to \E_1}\cdot \bfpi_{\E_1}&\cdots& \bff_{\I \to \E_1}\cdot \bfpi_{\E_n}&\0\\
\end{array} \right],
\label{eq:cesaro-convergence-P}
\end{multline}
in which $\bff_{\I \to \E_\ell}$ is a column vector of hitting probability from each state in $\I$ to class $\E_\ell$.

We note that while the Ces\`aro mean converges, the $t$-step transition probability $P^t$ does not necessarily converge. Consider, for example, the probability transition matrix $P = [\begin{smallmatrix} 0&1\\ 1&0 \end{smallmatrix}]$. In this case, $p_{12}^{(t)} = 1$ when $t$ is odd and $p_{12}^{(t)} = 0$ when $t$ is even, \ie $p_{12}^{(t)}$ is oscillating and never converges. In general, $P^t$ converges if and only if each of the closed communicating classes $\E_\ell$ for $\ell=1, \dots, n$ is \emph{aperiodic}. 
%More details are referred to the textbooks and are omitted here.

\section{Guaranteed outlier detection}\label{sec:prf-main}
%: proofs and discussions}
%\label{sec:prf-main}

Our outlier detection method by representation graph is guaranteed to correctly identify outliers in a union of subspaces when the representation is subspace-preserving and that the connectivity assumptions are satisfied. In this section, we first prove that the inliers and outliers in the data correspond to essential and inessential states, respectively, of the Markov chain associated with the representation graph (Lemma~\ref{thm:inlier-essential-equivalence}). Then, we show that the average of the first $T$ t-step probability distributions $\frac{1}{T} \sum_{t=1}^{T} \bfpi_0 P^t$ identifies essential and inessential states (Lemma~\ref{thm:cesaro-mean-converge}), thus establishing the correctness of our method.

\subsection{Proof of Lemma~\ref{thm:inlier-essential-equivalence}}

Recall that we work with a Markov chain with state space $\Omega=\{1, \cdots, N\}$, in which each state $i$ corresponds to the point $\x_i$ in the data matrix $X$.

First, we show that any inlier point $\x_i$ corresponds to an essential state of the Markov chain. Let $\x_j$ be any point such that $i \to j$. Since the representation matrix is subspace-preserving, we know that $\x_i$ and $\x_j$ lie in the same subspace. Furthermore, by Assumption~\ref{asmp:inlier}, all points in the same subspace are strongly connected, which implies that $j \to i$. Thus, $i$ is an essential state.

Second, we show that any outlier point $\x_i$ corresponds to an inessential state of the Markov chain. Consider the set $\Omega_i = \{k: i \to k\}$, \ie the set of points that are accessible from $\x_i$. By Assumption~\ref{asmp:outlier}, the set $\Omega_i$ cannot contain only outliers. Thus,  there exists $\x_j$ such that $i\to j$ and $\x_j$ is an inlier. However, since the representation is subspace-preserving, we know that $j \not\to i$. Therefore, $i$ is not an essential state, \ie it is an inessential state.

\subsection{Proof of Lemma~\ref{thm:cesaro-mean-converge}}

According to Theorem~\ref{thm:decomposition}, the state space of the Markov chain can be decomposed into $\I \cup \E_1 \cup \cdots \cup \E_n$, in which $\I$ contains the inessential states and each $\E_\ell$ is a closed communicating class containing essential states. Assume, without loss of generality, that the transition probability matrix has the form of \eqref{eq:decomposition-P}. By using \eqref{eq:cesaro-convergence-P}, the Ces\`aro mean in \eqref{eq:def-I} has the following limiting behavior:
\begin{multline}
\bfpi := \lim_{T \to \infty} \bar{\bfpi}^{(T)} =\lim_{T \to \infty} \frac{1}{T} \sum_{t=1}^{T} \bfpi_0 P^t \\
= \left[\frac{N_1 + \sum \bff_{\I \to \E_1}}{N} \cdot \bfpi_{\E_1}, \cdots, \frac{N_n + \sum \bff_{\I \to \E_n}}{N} \cdot \bfpi_{\E_n}, \0\right],
\label{eq:cesaro-limit}
\end{multline}
where $N_\ell$ for $\ell=1,\dots, n$ is the number of states in class $\E_\ell$, each $\bff_{\I \to \E_\ell}$ is a  vector of hitting probabilities for each state in $\I$ to class $\E_\ell$, and $\bfmu_{\E_\ell}$ is a positive vector of the stationary distributions of states in $\E_\ell$. Therefore, $\pi_j$ is zero if and only if $j$ is an inessential state. This finishes the proof.

\begin{table*}[t]
	\centering
	\small
	\label{tbl:time_EYaleB}
	\caption{Running time of experiments on Extended Yale B data with three inlier groups and $15\%$ outliers}
	\begin{tabular}{c||c|c|c|c|c|c|c|c}
		& OutRank &  CoP  & REAPER & OutlierPursuit &  LRR  & DPCP& $\ell_1$-thresholding &           R-graph (ours)            \\ \hline
		%	\multicolumn{8}{l}{\textsl{Inliers: all images from \textbf{one} subject ~~ Outliers: $35\%$, taken from other subjects }}    \\ \hline
		%Time (sec.) &  0.034  & 0.012 & 0.041  &     0.138      & 2.812 &         0.228         &           {0.106}            \\ \hline
		%\multicolumn{8}{l}{\textsl{Inliers: all images from \textbf{three} subjects ~~ Outliers: $15\%$, taken from other subjects }} \\ \hline
		Time (sec.) &  0.019  & 0.003 & 0.079  &     1.186      & 3.502 &  0.182&       0.312         &           {0.272}            \\ \hline
	\end{tabular}
\end{table*}

\subsection{Discussion}

In this section, we provide additional comments on using the Ces\`aro mean $\bar{\bfpi}^{(T)}$ in \eqref{eq:def-I} for outlier detection.

\myparagraph{Stationary distributions} By \eqref{eq:cesaro-limit}, the vector that $\bar{\bfpi}^{(T)}$ converges to is a stationary distribution of the Markov chain (see \eqref{eq:stationary-distribution-decomposition}). In fact, any convex combination of the stationary distribution of each closed communicating class is a stationary distribution of the Markov chain, and the particular stationary distribution that $\bar{\bfpi}^{(T)}$ converges to depends on the choice of the initial state distribution $\bfpi_0$. 

\myparagraph{A $T$-step probability distribution and PageRank} Traditionally, PageRank and many other spectral ranking algorithms use the limit of the $T$-step probability distribution $\bfpi^{(T)}$ rather than $\bar{\bfpi}^{(T)}$ as adopted in our method. However, the sequence $\bfpi^{(T)}$ converges if and only if each closed communicating class of the Markov chain is aperiodic, which is not necessarily satisfied in many cases. To address this, PageRank adopts a random walk with restart algorithm. It can be interpreted as a random walk on a transformed Markov chain that adds a small probability of transition from each state to the other states on the transition probability of the original Markov chain. By doing so, the transformed Markov chain contains a single communicating class that is aperiodic. Therefore, the stationary distribution necessarily becomes unique, and the sequence $\bfpi^{(T)}$ for the transformed Markov chain converges to the unique stationary distribution regardless of the initial state distribution. 

Despite the advantages of the random walk used by PageRank, all states of the Markov Chain are essential, so that outliers do not converge to zero probabilities. Therefore, it is less clear whether the stationary distribution that the algorithm converges to can effectively identify outliers. 

\section{Additional experimental results}
\label{sec:additional-experiment}

\subsection{Computational time comparison}

Table \ref{tbl:time_EYaleB} reports the average running time of the experiment on the Extended Yale B database with three inlier groups and $15\%$ outliers (226 images in total). From the table we observe that the running times of OutRank and CoP are much smaller than the other methods. This comes from the fact that OutRank and CoP are based on computing data pairwise inner products, which is efficient for small scale data. In contrast, the other methods solve optimization problems. In particular, REAPER, OutlierPursuit and LRR require computing an eigendecomposition of a matrix of size $D \times D$ ($D$ is the ambient dimension) during each iteration, which is time consuming when $D$ is large. In our experiments we observe that REAPER converges much faster than OutlierPursuit and LRR, thus the running time of REAPER is typically much smaller. The $\ell_1$-thresholding method and R-graph method (our algorithm) both compute the representation matrix by solving an $\ell_1$ optimization problem for each of the data points with all other data points as the dictionary. Subsequently, $\ell_1$-thresholding rejects outliers simply by computing the $\ell_1$ norms of the representations, while R-graph requires a random walk on the graph defined from the representation. We note that the random walk for R-graph is computationally efficient because of the sparsity of the representation matrix. In each step of the random walk, the computational complexity is on the order of $sN$ where $N$ is the number of data points and $s \ll N$ is the average number of nonzero entries in the representation vectors $\{\r_j\}$.

\begin{figure*}[t]
	\centering
	\subfigure[The effect of varying the parameter $\alpha$.\label{fig:varying_alpha}]{\includegraphics[scale=0.42]{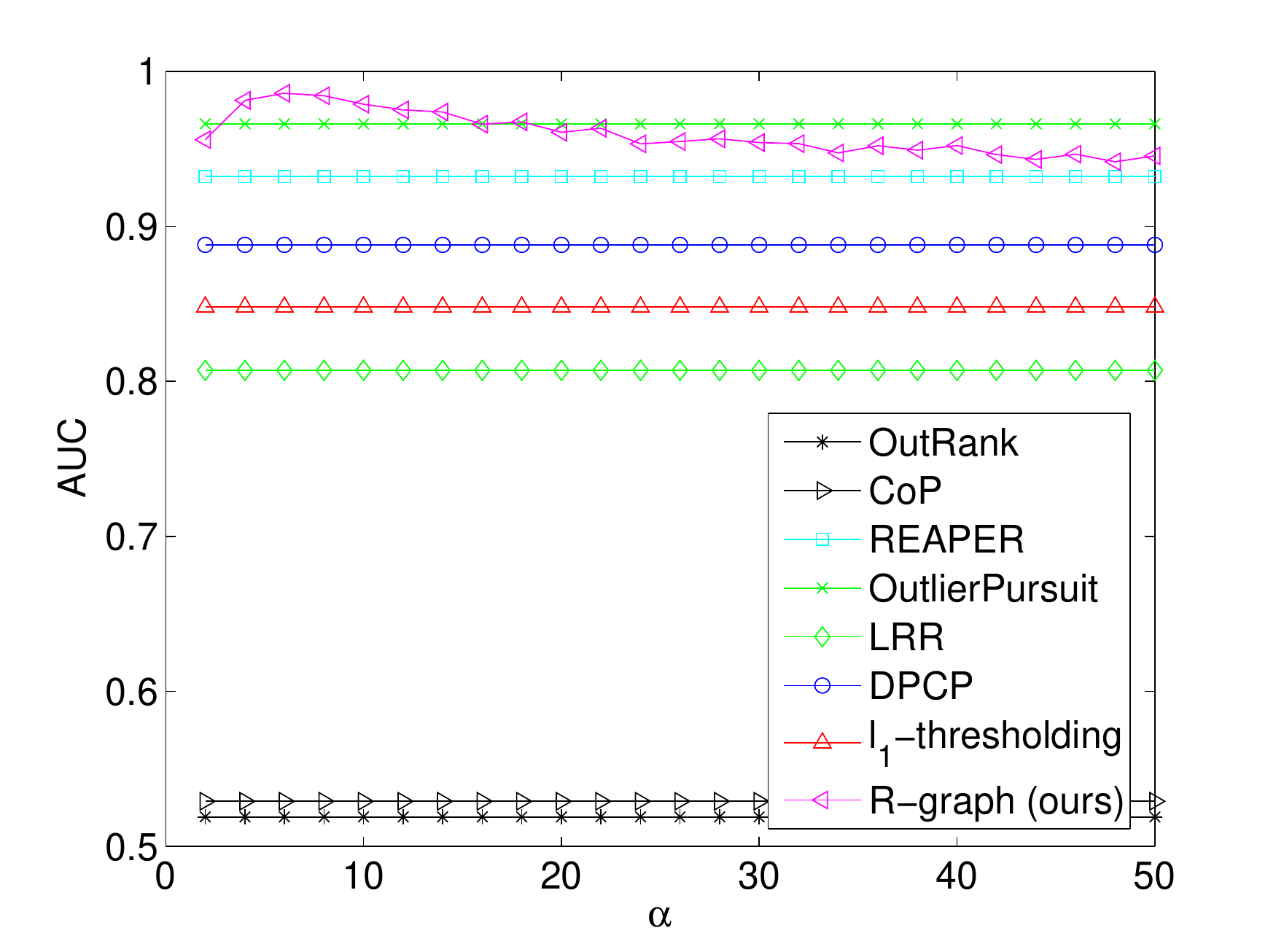}}
	\subfigure[The effect of varying the outlier percentage.\label{fig:varying_outliers}]{\includegraphics[scale=0.42]{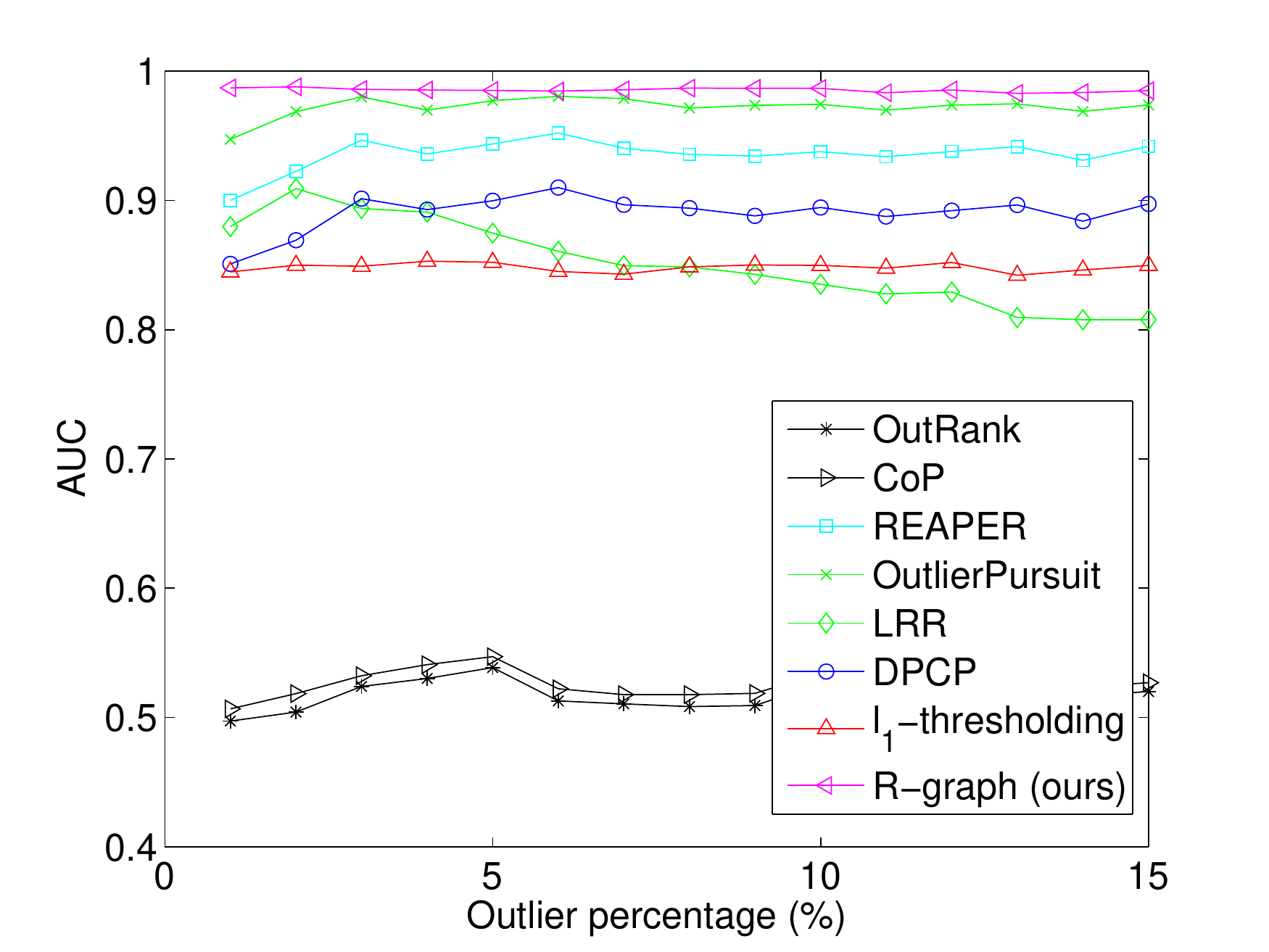}}
	\caption{Additional results for experiments on Extended Yale B with three inlier groups and $15\%$ outliers.}
\end{figure*} 

\begin{figure*}[t]
	\centering
	\subfigure[Inliers: $64$ images of one individual (displaying $10$ out of $64$).]{\includegraphics[scale = 0.5,trim={0cm 1.5cm  0cm 0.8cm},clip]{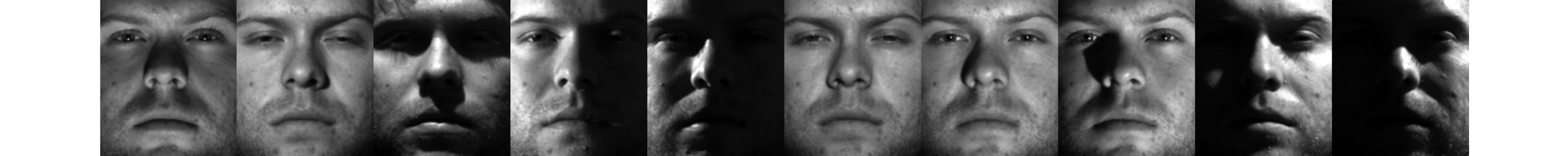}}
	\\
	\subfigure[Outliers: $10$ images from $10$ other individuals.]{\includegraphics[scale = 0.5,trim={0cm 1.5cm  0cm 0.8cm},clip]{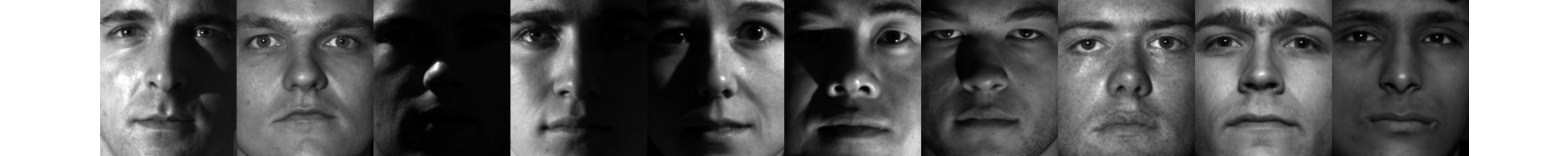}}
	\caption{An outlier detection dataset for visualizing the top $10$ outliers returned by diffferent methods.}
	\label{fig:EYaleB_example_data}
\end{figure*} 

\subsection{Influence of the algorithm parameters}

The first step of our method is to compute the data self-representation matrix using the optimization problem \eqref{eq:representation_matrix}. In this section, we illustrate the effect that the parameter $\gamma$ in \eqref{eq:representation_matrix} has on the performance of our method. Recall that for our numerical experiments we set $\gamma= \alpha \cdot \frac{\lambda}{\max_{i:i\ne j}|\x_j^\transpose \x_i|}$ and that the solution to \eqref{eq:representation_matrix} is nonzero if and only if $\alpha > 1$. We run experiments on Extended Yale B database with $3$ inlier groups and $15\%$ outliers while varying $\alpha$ in the range $[1, 50]$; the results are shown in Figure \ref{fig:varying_alpha}. We can see that  the R-graph performs well over a wide range of the parameter $\alpha$. For comparison, Figure \ref{fig:varying_alpha} also plots the performance of the other methods on the same dataset. 

\subsection{Influence of the percentage of outliers}

In this experiment, we fix the number of inlier groups to be $3$ and vary the percentage of outliers from $1\%$ to $15\%$. The performances of the different methods are reported in Figure \ref{fig:varying_outliers}. Note that the parameters for all methods are fixed across the different percentages of outliers. We see that the performance of our method is stable with respect to the percentage of outliers. Moreover, our method also achieves the best performance among all methods.

\subsection{Visualization of the outliers}
\begin{figure*}[h]
	\centering	
	\subfigure[Top 10 outliers by OutRank]{\includegraphics[scale = 0.88,trim={0cm 0cm  0cm 0cm},clip]{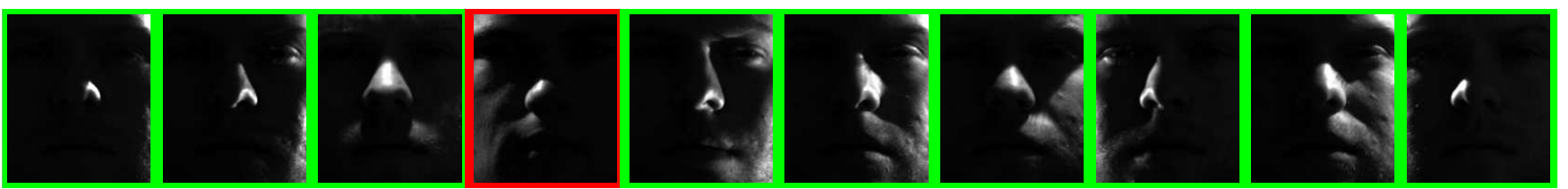}}
	\\
	\subfigure[Top 10 outliers by CoP]{\includegraphics[scale = 0.88,trim={0cm 0cm  0cm 0.1cm},clip]{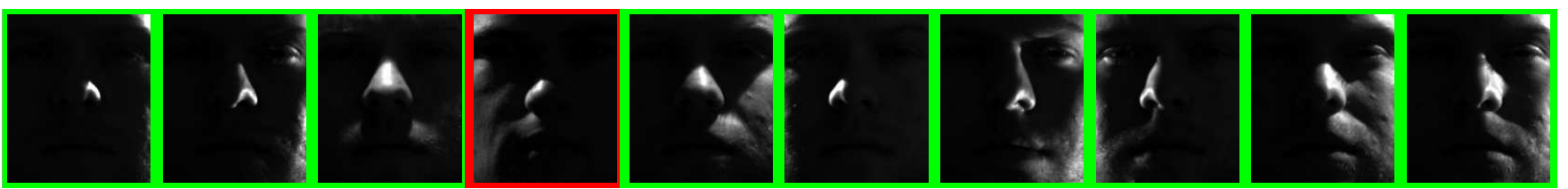}}
	\\
	\subfigure[Top 10 outliers by REAPER]{\includegraphics[scale = 0.88,trim={0cm 0cm  0cm 0cm},clip]{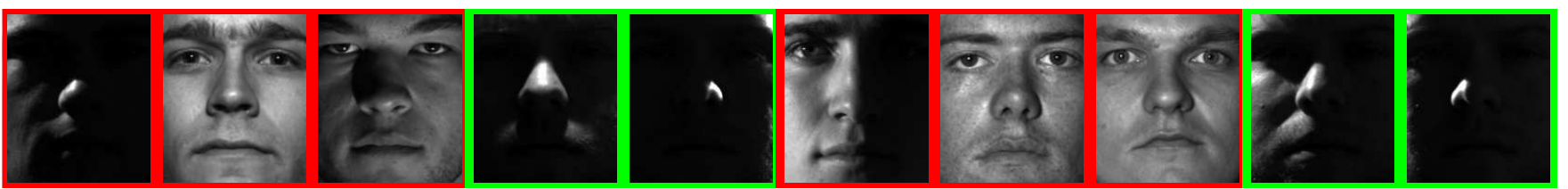}}
	\\
	\subfigure[Top 10 outliers by OutlierPursuit]{\includegraphics[scale = 0.88,trim={0cm 0cm  0cm 0cm},clip]{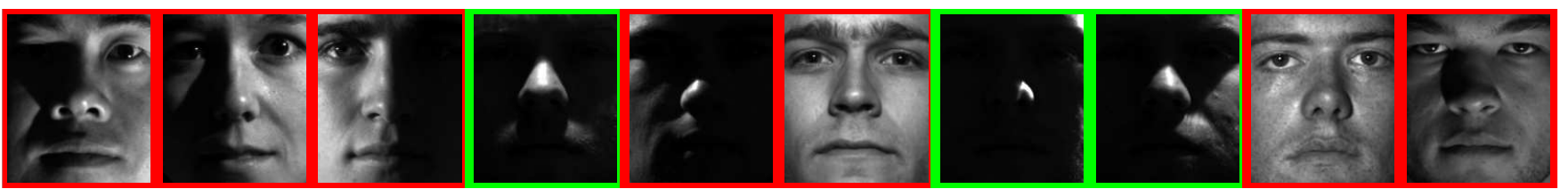}}
	\\
	\subfigure[Top 10 outliers by LRR]{\includegraphics[scale = 0.88,trim={0cm 0cm  0cm 0cm},clip]{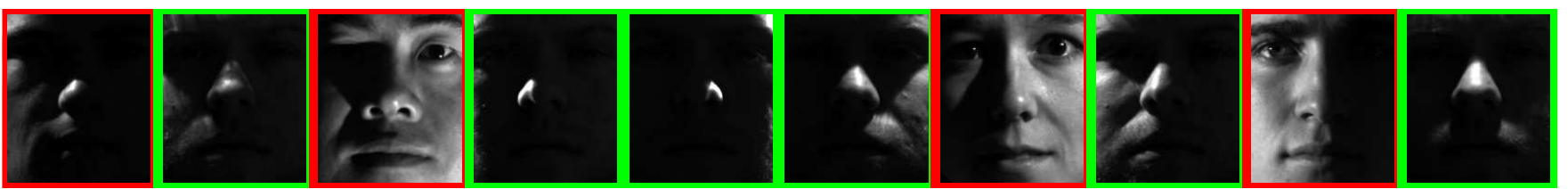}}
	\\
	\subfigure[Top 10 outliers by DPCP]{\includegraphics[scale = 0.88,trim={0cm 0cm  0cm 0cm},clip]{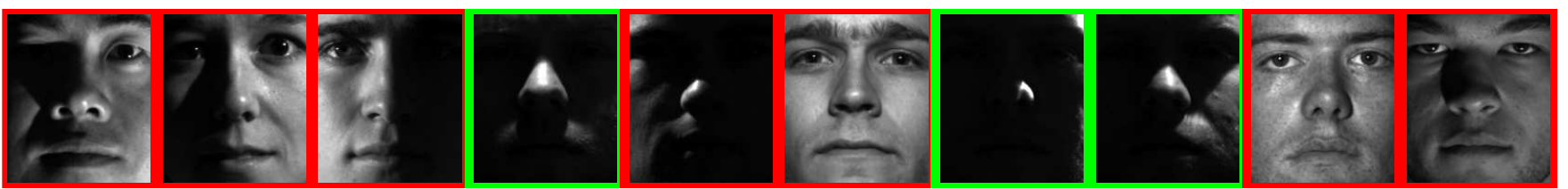}}
	\\
	\subfigure[Top 10 outliers by $\ell_1$-thresholding]{\includegraphics[scale = 0.88,trim={0cm 0cm  0cm 0cm},clip]{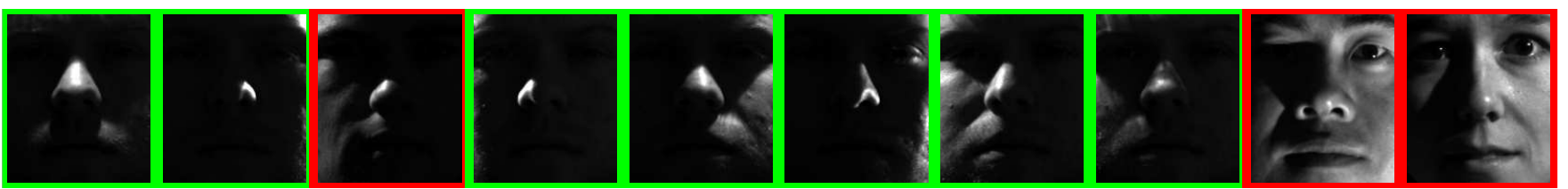}}
	\\
	\subfigure[Top 10 outliers by R-graph (ours)]{\includegraphics[scale = 0.88,trim={0cm 0cm  0cm 0cm},clip]{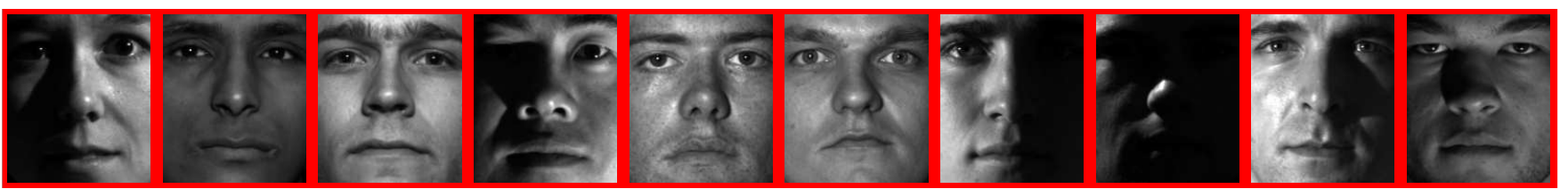}}
	\caption{Visualizing the top $10$ outliers from different methods. Image in red box: true outlier. Image in green box: true inlier.}
	\label{fig:EYaleB_example_result}
\end{figure*}

To supplement the AUC and F1 measures previously provided, and also to better understand the outliers returned by our outlier detection method, we conducted additional experiments that display the top outliers detected in each experiment. 
	%We construct a dataset with $10$ outliers so that it is easy to display all $10$ outliers. 
The set of inliers is taken to be the $64$ images of the first subject of the Extended Yale B database, and the outlier set is chosen as $10$ images randomly chosen from the remaining $37$ subjects (see Figure~\ref{fig:EYaleB_example_data}). The top $10$ outliers returned by different methods are reported in Figure~\ref{fig:EYaleB_example_result}. Images with red boxes are outliers (\ie true positives) and images with green boxes are inliers (\ie false positives).

False positives for all methods are mostly images taken under extreme illumination conditions. Such images have large shadows, which has the effect of removing them from the underlying subspace associated with the individual thus making them more likely to be detected as outliers. The results show that REAPER, Outlier Pursuit, DPCP and R-graph are relatively robust.  In particular, R-graph is significantly better than $\ell_1$-thresholding even though both are sparse representation based methods. This shows that while the magnitude of the representation vector adopted by $\ell_1$-thresholding can be sensitive to corruptions, the connectivity behavior explored by R-graph is more robust.

\end{appendices}

{\small
	\bibliographystyle{ieee}
	\bibliography{biblio/vidal,biblio/vision,biblio/math,biblio/learning,biblio/sparse,biblio/geometry,biblio/dti,biblio/recognition,biblio/surgery,biblio/dataset,biblio/matrixcompletion,biblio/segmentation}
}

\end{document}